\PassOptionsToPackage{table}{xcolor}
\documentclass{article} 
\usepackage{iclr2027_conference,times}

\usepackage{amsmath,amsfonts,bm}

\def\eqref#1{equation~\ref{#1}}

\def\1{\bm{1}}

\DeclareMathAlphabet{\mathsfit}{\encodingdefault}{\sfdefault}{m}{sl}
\SetMathAlphabet{\mathsfit}{bold}{\encodingdefault}{\sfdefault}{bx}{n}

\usepackage{hyperref}
\usepackage{url}
\usepackage{graphicx}
\usepackage{adjustbox}
\usepackage{fontawesome5}

\usepackage{algorithm}
\usepackage{algpseudocode}
\usepackage{amssymb}
\usepackage{booktabs}
\usepackage{multirow}
\usepackage{tabularx}
\usepackage{array}
\usepackage[table]{xcolor}
\usepackage[textsize=small]{todonotes}
\usepackage{titletoc}

\usepackage{fvextra}
\usepackage[most]{tcolorbox}
\usepackage{wrapfig}
\usepackage{needspace}
\usepackage{subcaption}

\fvset{
  breaklines=true,
  breakanywhere=true,
  breaksymbolleft={},   
  breaksymbolright={},  
  breakindent=0pt,
  breakautoindent=false,
  fontsize=\footnotesize,
  frame=none,
  tabsize=4
}

\newtcolorbox{promptbox}[1]{
  enhanced,
  breakable,
  colback=black!2,
  colframe=black!30,
  boxrule=0.4pt,
  arc=1mm,
  left=2.5mm,
  right=2.5mm,
  top=2.5mm,
  bottom=2.5mm,
  colbacktitle=black!7,
  coltitle=black,
  fonttitle=\small\bfseries,
  title={#1},
  title after break={#1\space(continued)},
  before skip=8pt,
  after skip=12pt
}

\def\stripext#1.txt\relax{#1}

\definecolor{citecolor}{HTML}{377fbd}
\definecolor{linkcolor}{HTML}{c0392b}
\hypersetup{
    colorlinks=true,
    linkcolor=linkcolor,
    citecolor=citecolor,     
    urlcolor=black,
}

\newcommand{\ModelName}{WebCoder}

\newcommand{\urlHomepage}{https://douzj-llm.github.io/WebCoder/}
\newcommand{\urlCode}{https://github.com/douzj-llm/WebCoder}
\newcommand{\urlModel}{https://huggingface.co/Roy-DOU/WebCoder-27B}
\newcommand{\urlDataset}{https://huggingface.co/datasets/Roy-DOU/Webdev-Bench}
\definecolor{resourcecolor}{HTML}{377fbd}
\newcommand{\resourcelink}[3]{\href{#1}{\textcolor{resourcecolor}{#2\,#3}}}
\newcommand{\resourcelinks}{%
  \resourcelink{\urlHomepage}{\faGlobe}{Homepage}\hspace{2.1em}%
  \resourcelink{\urlCode}{\faGithub}{Code}\hspace{2.1em}%
  \resourcelink{\urlModel}{\faRobot}{Model}\hspace{2.1em}%
  \resourcelink{\urlDataset}{\faDatabase}{Dataset}}

\definecolor{rankSecondColor}{RGB}{0,92,153}
\definecolor{rankBestColor}{RGB}{192,7,4}
\newcommand{\rankbest}[1]{\textcolor{rankBestColor}{\textbf{#1}}}
\newcommand{\ranksecond}[1]{\textcolor{rankSecondColor}{\underline{#1}}}

\title{\raisebox{-0.1ex}{\adjustbox{height=1.8em, valign=m}{\includegraphics[height=2.5em]{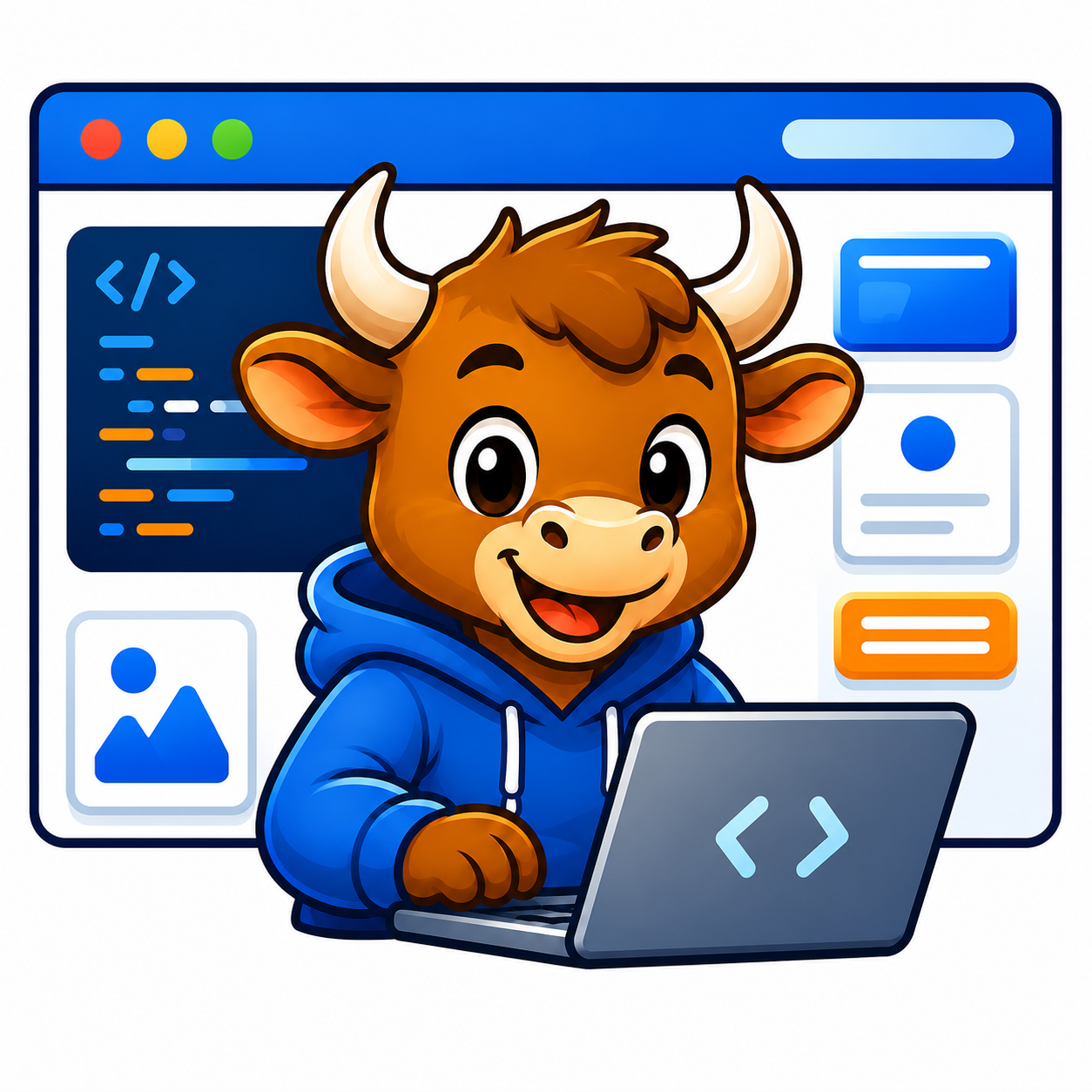}}}\hspace{0.25em}\ModelName{}: Scaling General Frontend Development with Artifact-Grounded Agentic Reinforcement Learning}

\author{%
\bfseries Zijun Dou$^{1,2*}$, Zhen Yang$^{1,2*\dagger}$, Xijun Liu$^{2,3}$, Dongzhuoran Zhou$^{4}$,\\
\bfseries Aohan Zeng$^{1,2}$, Jie Tang$^{1\dagger}$\\
\normalfont\mdseries $^{1}$Tsinghua University, $^{2}$Zhipu AI, $^{3}$Beihang University, $^{4}$University of Oslo\\
\normalfont\mdseries\texttt{royz.dou@outlook.com,\ yang-zhen@mail.tsinghua.edu.cn}\\
\normalfont\mdseries\texttt{jietang@tsinghua.edu.cn}\\[1.2ex]
\normalfont\mdseries\makebox[\dimexpr\textwidth-2\tabcolsep\relax][c]{\resourcelinks}%
}

\newcommand{\blfootnote}[1]{%
  \begingroup
  \renewcommand\thefootnote{}\footnote{#1}%
  \addtocounter{footnote}{-1}%
  \endgroup
}

\iclrfinalcopy 
\begin{document}

\maketitle
\lhead{Under review}  
\blfootnote{$^{*}$\,Core contributors; $^{\dagger}$\,Corresponding authors.\\
\hspace*{1.1em}Work was done when ZD, ZY, XL, AZ interned at Zhipu AI.}

\begin{abstract}

Real-world frontend development requires agents to compose reproduction, generation, modification, and repair over an evolving application state. Learning a general policy for this process requires translating heterogeneous execution feedback into task-aligned rewards and effective credit assignment across long sequences of interdependent actions. 
We introduce \textbf{\ModelName{}}, a general frontend development agent trained with artifact-grounded agentic reinforcement learning. 
Our key idea is to assess diverse development activities through their effects on a shared executable artifact and its satisfaction of task requirements. 
We build Webdev-World, an executable environment that integrates code editing, application execution, browser inspection, and on-demand skills while preserving project state across interactions. 
An independent verifier evaluates the final application against task requirements to provide terminal rewards, while changes in intermediate artifact progress redistribute the resulting rollout advantage across development segments. 
Context compaction and message-tree sample construction support training on trajectories beyond a single context window while avoiding repeated loss computation on shared generated history. 
We further introduce Webdev-Bench to evaluate composed development requirements within a single project. On Webdev-Bench, \ModelName{}-27B raises requirement satisfaction from 64.8 for its Qwen3.8-27B base model to 72.8, exceeding Claude-Opus-4.8 (65.1) and GPT-5.4 (59.6); on Vision2Web, it improves the average visual and functional score from 41.1 to 62.4, on par with Claude-Opus-4.8. Ablation studies confirm the contributions of artifact-grounded credit assignment and long-horizon training.

\end{abstract}

\section{Introduction}







\begin{figure}[h]
\centering
\includegraphics[width=1.0\textwidth]{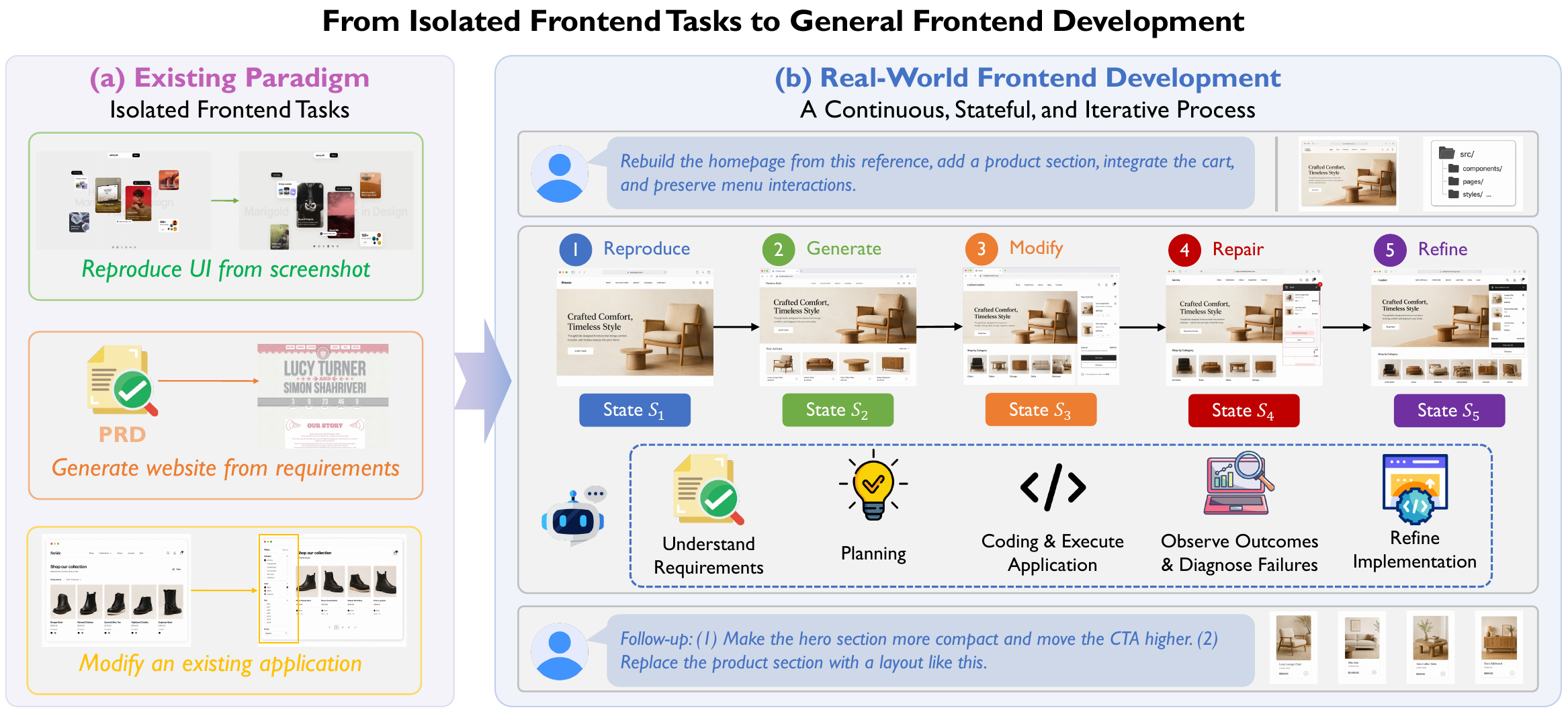}
\caption{\textbf{From isolated frontend tasks to general frontend development.}
(a) Existing work typically studies frontend reproduction, generation, and modification as separate task categories.
(b) In real-world development, a single user request often requires multiple capabilities to be composed within a continuous, stateful, and iterative development process, followed by multi-turn and multimodal refinement over the evolving application state.}
\vspace{-6mm}
\label{fig:intro}
\end{figure}

Recent advances in multimodal code generation have expanded the capabilities of frontend agents, from reproducing webpage designs~\citep{si2025design2code,xu2025webvia,xu2026video2code} to refining interfaces through rendered feedback~\citep{yang2026ui2code,deng2026visrefiner} and generating interactive websites~\citep{lu2026webgen-agent}. In practice, these capabilities often need to work together within the same application. A user may ask an agent to reproduce a reference design, add a product section, connect it to an existing shopping cart, and preserve the navigation behavior. These requirements share code and application state: implementing a new feature can affect the layout, while a visual revision can disrupt an existing interaction. As illustrated in Figure~\ref{fig:intro}, completing such a request requires reproduction, generation, modification, and repair to be coordinated throughout development. This raises a central question: \emph{how can we train a general frontend agent to compose these capabilities over an evolving application state?}

Training such an agent requires coordinating activities whose local goals differ but whose effects are coupled through shared code and application state. Satisfying one requirement can violate another, making it necessary to assess the combined effects of these activities on the application. Progress may become observable only after several edits and executions, while later changes can introduce regressions, making it difficult to distinguish the contributions of different development stages. The interaction history can also outgrow the model's context window as the project evolves. Therefore, learning a general policy requires a common basis for evaluating heterogeneous activities, credit assignment across interdependent stages, and support for long-horizon training.


To address these challenges, we introduce \textbf{\ModelName{}}, a general frontend development agent trained with \emph{artifact-grounded agentic reinforcement learning}. We unify heterogeneous frontend tasks within \textbf{Webdev-World}, an executable environment that provides a shared, task-agnostic interface for code editing, execution, browser inspection, and on-demand skills while preserving project state across interactions. This common interaction space allows a single policy to interleave reproduction, generation, modification, and repair as the application evolves. 
Our key idea is to ground policy learning in the evolution of an executable application, using task requirements to evaluate both the final outcome and intermediate development progress. 
Evaluating build results, rendered pages, and interaction outcomes against the same requirements provides a common basis for assessing the combined effects of development activities and optimizing the policy from its own trajectories.



Specifically, we construct a weighted checklist from the task requirements and use an independent verifier to build, run, and inspect the final application. Deterministic checks and visual assessments measure requirement satisfaction, with the aggregated score gated by successful execution and rendering to form the terminal reward. This terminal reward evaluates the final outcome but does not distinguish which development stages advanced the task or introduced regressions. Thus, we estimate intermediate requirement satisfaction from available project and execution evidence using the same checklist. Changes in these estimates provide stage-level progress and regression signals, which guide credit assignment across the trajectory and help the policy learn from the development process.
For training across context windows, we combine context compaction with message-tree sample construction. Compaction lets development continue while project state persists in Webdev-World. A message tree records histories across context rewrites, enabling training on different stages of the same trajectory. Shared generated history remains available as context but contributes to the policy loss at most once.


To evaluate capability composition, we introduce \textbf{Webdev-Bench}, which combines development requirements within a single project, including requested changes and the preservation of specified existing behavior. 
On Webdev-Bench, \ModelName{}-27B reaches a requirement satisfaction score of 72.8 against 64.8 for its Qwen3.8-27B base model and 65.1 for Claude-Opus-4.8, and delivers a running application on all 40 tasks. On Vision2Web~\citep{he2026vision2web}, it achieves an average visual and functional score of 62.4, compared with 41.1 for the base model and 62.5 for Claude-Opus-4.8.

Our main contributions are summarized as follows:
\begin{itemize}
    \item We study \textbf{general frontend development} as capability composition over a persistent project state, and introduce \textbf{Webdev-World} as a shared environment for training \textbf{\ModelName{}}, a general frontend development agent.

    \item We develop \textbf{artifact-grounded agentic reinforcement learning}, using task requirements to connect final application verification with progress-based credit assignment, and supporting training over trajectories that span multiple context windows.

    \item We introduce \textbf{Webdev-Bench} to assess capability composition and preservation of specified existing behavior, and demonstrate improved visual and functional performance on Vision2Web.

\end{itemize}

\section{Method}
\label{sec:method}

\begin{figure}[h]
\centering
\includegraphics[width=\textwidth]{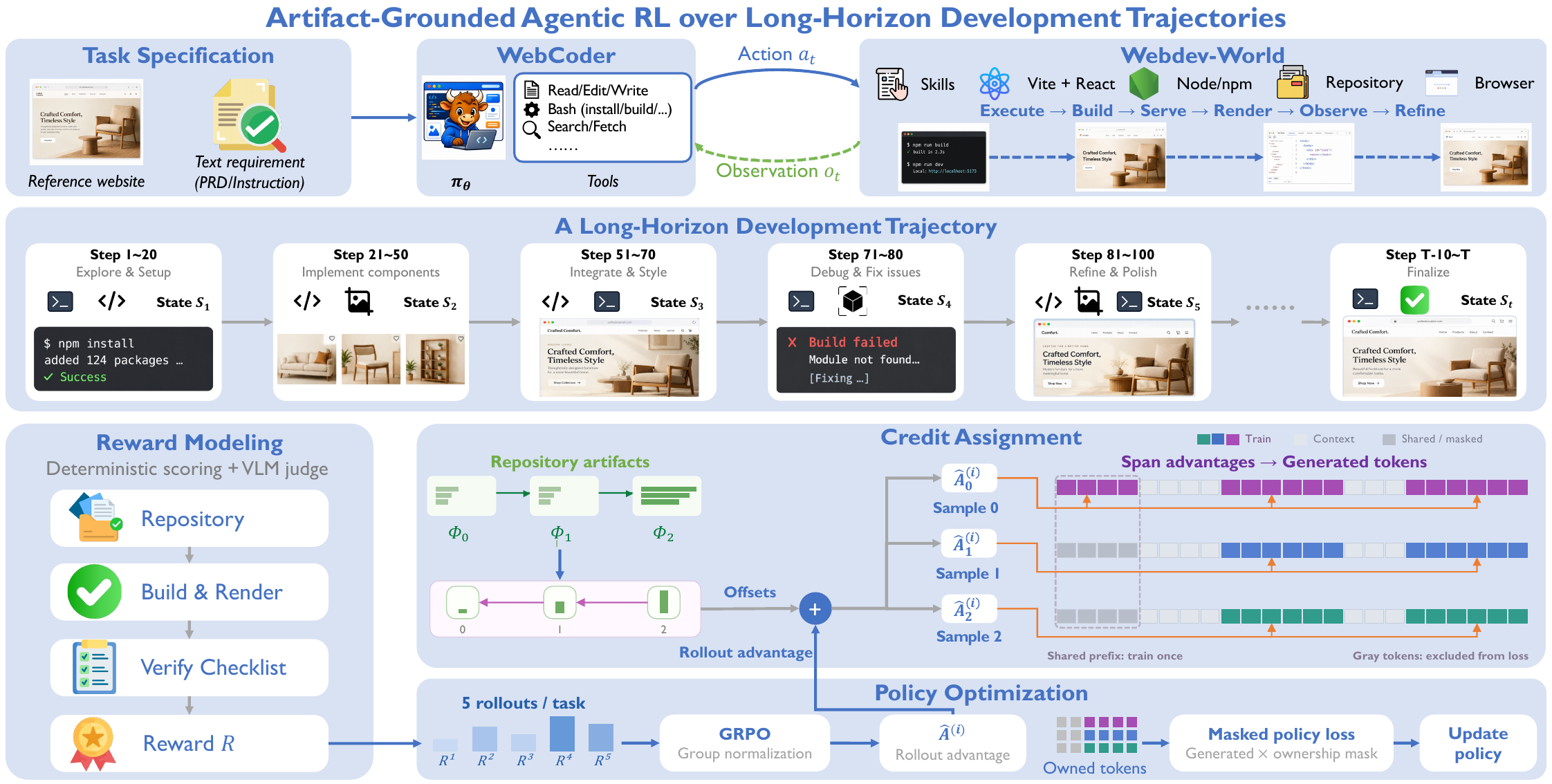}
\caption{\textbf{Overview of our artifact-grounded agentic reinforcement learning framework.}
\ModelName{} builds and refines a runnable project in Webdev-World through development tools and on-demand skills; an independent verifier scores the final application against the task checklist, artifact-based potentials redistribute the GRPO rollout advantage across trajectory segments, and an ownership mask assigns each generated token to one training sample.
}
\label{fig:method}

\end{figure}

We present an artifact-grounded agentic reinforcement learning framework for training \ModelName{} to compose frontend development capabilities over a persistent project state (Figure~\ref{fig:method}). We formulate the task, introduce Webdev-World as the shared environment, describe how one requirement decomposition connects terminal rewards with progress-based credit assignment, and detail training on trajectories that span multiple context windows.

\subsection{Problem Statement}

We study general frontend development as a sequential process in which a single agent coordinates reproduction, generation, modification, and repair within an evolving project. Given a task input \(x\), which may include reference screenshots, natural-language requirements, and an initial repository, the agent iteratively edits, executes, and inspects the application; success is determined by whether the final application jointly satisfies the task requirements, including constraints on existing behavior.

We model this process as a partially observable decision process. At time \(t\), the policy \(\pi_\theta\) generates an action \(a_t\) (text output and tool calls) conditioned on the context \(c_t=\mathcal{C}(h_t)\), where \(h_t=[p,x,a_0,o_0,\ldots,a_{t-1},o_{t-1}]\) is the interaction history under system instructions \(p\), \(o_t\) is the observation returned after executing \(a_t\) (command output, file contents, or a page screenshot), and \(\mathcal{C}\) is a context management mechanism that keeps the information needed for the task within a finite window. A trajectory \(\tau\) terminates at time \(T\) with a final project state \(\mathcal{A}_T(\tau)\). A verifier \(R\) independently builds, runs, and inspects the resulting application and returns a scalar terminal reward \(R(\mathcal{A}_T(\tau),x)\), using only evidence from the application and its execution rather than the agent's conversation or self-reported claims. We train \(\pi_\theta\) to maximize the expected terminal reward:
\begin{equation}
    \max_{\theta}\;
\mathbb{E}_{x\sim\mathcal{D}}
\mathbb{E}_{\tau\sim\pi_\theta(\cdot\mid x)}
\left[
R\bigl(\mathcal{A}_T(\tau),x\bigr)
\right],
\end{equation}
where \(\mathcal{D}\) denotes the task distribution.

\subsection{Webdev-World}

Webdev-World is an executable environment with a common interface for code editing, command execution, browser inspection, and on-demand skills. Each rollout runs in an isolated container initialized with a React/Vite scaffold or an existing project, and project files persist across interactions, so a single policy can interleave development activities over the same evolving application. The policy edits the repository through file and shell tools and invokes \texttt{webshot} to build and serve the application and obtain rendered screenshots and DOM statistics, which show how preceding changes affect the application; on-demand skills provide procedural guidance for page decomposition, asset preparation, code generation, validation, and repair.


\subsection{Artifact-Grounded Agentic Reinforcement Learning}
\label{sec:method-rl}

We ground policy learning in the application's satisfaction of task requirements, using final verification to compute terminal rewards and intermediate progress estimates to guide credit assignment.

\paragraph{Reward Modeling.}

Frontend tasks combine visual, content, functional, and interaction requirements that the final application must satisfy jointly. From the requirements, reference images, and project context in \(x\), we construct a weighted checklist \(\{(w_i,v_i)\}_{i=1}^{m}\), where \(w_i\) is the weight of requirement \(i\) and \(v_i\) is its verification function, chosen among deterministic checks for directly testable requirements, visual comparison against the reference for reproduction requirements, and rubric-guided judgments by a VLM (GLM-5V-Turbo~\citep{hong2026glm}) for quality requirements that resist deterministic verification. An independent verifier \(R\) builds, runs, and inspects the final project state \(\mathcal{A}_T\) under runtime conditions consistent with the development environment and aggregates the checklist as
\begin{equation}
Q(\mathcal{A}_T,x)
=
\sum_{i=1}^{m}w_i\,v_i(\mathcal{A}_T,x),
\qquad
v_i(\mathcal{A}_T,x)\in[0,1],
\quad
w_i\geq 0,
\quad
\sum_{i=1}^{m}w_i=1.    
\end{equation}

To ensure that the delivered application can be executed and inspected, we gate this quality score by execution validity. Let \(G(\mathcal{A}_T)\in\{0,1\}\) indicate whether the project passes all build, entry-point resolution, and independent browser-rendering checks. The terminal reward returned by the verifier is
\begin{equation}
    R(\mathcal{A}_T,x)
=
G(\mathcal{A}_T)\,Q(\mathcal{A}_T,x).
\end{equation}
Appendix~\ref{app:reward-credit} lists the evidence, verification method, and weight of each requirement type.


\paragraph{Credit Assignment.} The terminal reward measures task completion in the final project but does not directly distinguish the contributions of different stages in a long development trajectory. Thus, we estimate intermediate requirement satisfaction using the same checklist as final verification and use changes in these estimates to adjust stage-level advantages.
The effect of a code change becomes observable only after the application is built, run, and inspected, so progress cannot be attributed to individual tool calls. We therefore assign credit at the level of segments, taking context compaction boundaries as natural segment boundaries, and measure progress by the change in project state between consecutive boundaries.

To align progress measurement with task evaluation, we retain the checklist decomposition and estimate requirement satisfaction at segment boundaries.
Let \(\phi_{k,i}\in[0,1]\) denote the estimated satisfaction of requirement \(i\) at boundary \(k\). For tasks with a requirement checklist, \(\phi_{k,i}\) is the fraction of the requirement's expected content present in the most recent page the agent fetched from its own running application at or before boundary \(k\); the estimate is updated only by such fetches, carries the previous value forward when no new fetch occurred, and records any decrease in full when a new fetch shows regression. For reproduction tasks without a checklist, a structural authoring score over the source files present at boundary \(k\) takes its place (Appendix~\ref{app:reward-credit}).
We aggregate these estimates with the checklist weights into an artifact potential, whose difference across consecutive boundaries represents observable progress within a segment:
\begin{equation}
    \Phi_k=\sum_{i=1}^{m}w_i\,\phi_{k,i},
\qquad
\Delta\Phi_k=\Phi_k-\Phi_{k-1}.
\end{equation}
We combine this process signal with the rollout-level advantage derived from terminal rewards.
Let \(\widehat A_{\mathrm{rollout}}\) denote the GRPO advantage computed from terminal rewards for the same task. The segment adjustment \(b_k\) is the discounted return of potential differences,
\begin{equation}
    b_k=\sum_{j\ge k}\gamma^{\,j-k}\,r_j,
\qquad
r_j=\gamma\,\Phi_{j+1}-\Phi_j\ \ (j<K-1),
\qquad
r_{K-1}=0,
\end{equation}
with \(\gamma=0.995\); the last segment carries no shaped reward because the terminal reward already enters through \(\widehat A_{\mathrm{rollout}}\).
To redistribute credit across segments while preserving the mean advantage determined by terminal evaluation, we apply token-weighted centering and clipping:
\begin{equation}
    \delta_k
=
\operatorname{clip}\!\left(b_k-
\frac{\sum_{j\in\mathcal K}n_jb_j}
     {\sum_{j\in\mathcal K}n_j},\,\pm c\right),
\qquad
\widehat A_k
=
\widehat A_{\mathrm{rollout}}+\delta_k,
\end{equation}
where \(\mathcal K\) is the set of segments selected for training, \(n_k\) is the number of generated tokens in segment \(k\) that contribute to the policy loss, and \(c=5\); the offsets are re-centered after clipping so that \(\sum_{k\in\mathcal K}n_k\delta_k=0\) holds exactly. Process signals thus redistribute credit across segments, while the terminal reward remains the trajectory-level evaluation baseline.

Development segments define the units of credit assignment, while message-tree paths define the training samples: each retained generated token receives the advantage of the segment in which it was generated, and an ownership mask assigns it to exactly one training sample, so tool responses and repeated history serve only as context and shared history is never optimized twice.


\paragraph{Policy Optimization.} We optimize the policy with a masked, clipped objective that applies segment-level advantages to generated tokens in the constructed training samples.
For each task \(x\), we sample \(N\) trajectories from the behavior policy \(\pi_{\theta_{\mathrm{old}}}\) and obtain terminal rewards
\(r_j=R(\mathcal{A}_T(\tau^{(j)}),x)\).
Following GRPO~\citep{shao2024deepseekmath}, we compute the rollout-level advantage as
\begin{equation}
    \widehat A_{\mathrm{rollout}}^{(i)}
=
\frac{r_i-\mu_x}{\sigma_x+\epsilon},
\qquad
\mu_x=\frac{1}{N}\sum_{j=1}^{N} r_j,
\end{equation}
where \(\sigma_x\) is the standard deviation of rewards within the group and \(\epsilon\) is a numerical stability constant; credit assignment then yields segment-level advantages.


Let \(u\) index token positions in the training batch, \(y_u\) denote a generated token, and \(c_u\) denote its generation context. The probability ratio between the current and behavior policies is
\begin{equation}
    \rho_u(\theta)
=
\frac{\pi_\theta(y_u\mid c_u)}
{\pi_{\theta_{\mathrm{old}}}(y_u\mid c_u)}.
\end{equation}
Let \(\widehat A_u\) be the advantage of the segment containing token \(u\), and let \(M_u\in\{0,1\}\) be the joint generation and ownership mask. The policy loss is
\begin{equation}
    \mathcal L_{\mathrm{policy}}(\theta)
=
-\frac{1}{\sum_u M_u}
\sum_u M_u
\min\!\left(
\rho_u(\theta)\widehat A_u,\,
\operatorname{clip}\!\left(\rho_u(\theta),1-\eta,1+\eta\right)
\widehat A_u
\right),
\end{equation}
where \(\eta\) is the clipping threshold.


\paragraph{Training Beyond the Context Window.}
To support trajectories beyond the context window, we apply tool-output truncation, tool-result folding, and summary compaction.
After compaction, the agent continues from the summarized history while project state persists in the environment.

For training, we record conversation histories in a message tree that preserves branches across context rewrites. Root-to-leaf paths form candidate training samples. Each rollout retains its initial and terminal candidates and a bounded number of additional samples selected by artifact progress, falling back to evenly spaced selection when measurements are unavailable. 
Ownership masks based on message position and content ensure that repeated generated tokens contribute to the loss at most once. 
Samples receive the segment advantages defined above, while each rollout contributes only one terminal reward to GRPO group statistics. 
A path exceeding the 65{,}536-token training budget keeps its most recent tokens so that the trainable segment is preserved, and the behavior-policy log-probabilities in \(\rho_u\) are recomputed on the assembled sequence (Appendix~\ref{app:reward-credit}).
The algorithm is detailed in Appendix~\ref{sec:context-samples}.
Rollout infrastructure, including speculative decoding, prefix caching, and log-probability monitoring between rollout and training engines, is described in Appendix~\ref{sec:rollout-infra}.


\section{Experiments}

\subsection{Experimental Setup}
\label{sec:setup}

\ModelName{}-27B is initialized from Qwen3.8-27B and trained with GRPO in Webdev-World, with 16 tasks per step, 8 rollouts per task, a learning rate of $2\times10^{-6}$, a 116{,}736-token rollout context, and a 65{,}536-token training context. Training proceeds in two stages: the first uses single-type tasks (text-to-project generation and single- and long-screenshot replication), and the second adds composite tasks that replicate a page from scratch and then extend and modify it. All hyperparameters are listed in Table~\ref{tab:training-config}.

We evaluate on two benchmarks. \textbf{Vision2Web} (L2-frontend, 66 titles) provides reference screenshots of a multi-page site and asks for a runnable frontend project; its official scorer reports a visual score (VS), a functional test pass rate (FS), and their average (Avg). \textbf{Webdev-Bench} is our benchmark of 40 composite tasks (188 requirements, 1{,}125 hidden checklist items) that start from a reference screenshot, a textual specification, or an existing repository; requirements are revealed and gated round by round, and an independent acceptance agent operates the deliverable and grades every item, giving the requirement satisfaction score (RS) and its visual, functional, and interaction components (VS, FS, IS). Both benchmarks also report the delivery success rate (DSR), and undelivered tasks count as $0$; definitions and protocols are given in Appendix~\ref{sec:webdev-bench}.

We compare against six proprietary models (Claude-Opus-4.8~\citep{anthropic2026opus48}, GPT-5.4~\citep{openai2026gpt54}, GPT-5~\citep{openai2025gpt5}, Gemini-3-Pro~\citep{google2025gemini3pro}, Gemini-3-Flash~\citep{google2025gemini3flash}, Seed-1.8~\citep{seed2026seed18}) and eight open-source models (Kimi-K2.5~\citep{team2026kimi25}, MiMo-V2.5~\citep{xiaomi2026mimo25}, Step-3.7-Flash~\citep{stepfun2026step37flash}, Qwen3.8-27B~\citep{qwen2026qwen38}, Qwen3.5-397B-A17B and Qwen3.5-122B-A10B~\citep{qwen2026qwen35}, Qwen3-VL-32B/8B-Instruct~\citep{yang2025qwen3}); see Appendix~\ref{app:baselines}.
All models are evaluated with the same agent framework, tools, prompts, container environment, time budget, and judges.
Judge consistency is analyzed in Appendix~\ref{app:verifier}.

\subsection{Main Results}
\label{sec:main-results}

Table~\ref{tab:main_results} reports the main results. On Webdev-Bench, \ModelName{}-27B reaches an RS of 72.8, above the base model Qwen3.8-27B (64.8) and the strongest baseline Claude-Opus-4.8 (65.1); the remaining proprietary models lie between 48.0 and 59.6. All three components are also the highest: VS 81.9, FS 73.8, and IS 90.8, where the interaction component leads the runner-up (Gemini-3-Flash, 85.0) by 5.8 points, indicating that the gains come mainly from interactive behavior that works correctly after multi-round development. \ModelName{} delivers all 40 tasks, against 95\% for the base model. Larger open-source models do not translate scale into results: Qwen3.5-397B-A17B and Qwen3.5-122B-A10B reach RS 28.7 and 24.6, and the two Qwen3-VL-Instruct models rarely complete a task within the agent framework. Appendix~\ref{app:full-results} splits the results by task starting point: the gains of \ModelName{} concentrate on text-to-project tasks (RS 70.0 versus 57.2 for the base model) and replication tasks (58.9 versus 51.7).

On Vision2Web, \ModelName{}-27B reaches an Avg of 62.4, on par with Claude-Opus-4.8 (62.5) and above GPT-5.4 (57.9) and all open-source models; its FS of 67.1 is the highest (tied with Qwen3.5-397B-A17B) and its VS of 57.7 is second to Claude-Opus-4.8's 59.1. Relative to the base model, Avg improves from 41.1 by 21.3 points, VS and FS improve by similar margins, and the delivery rate rises from 48.5\% to 80.3\%.

\begin{table}[h]
    \centering
    \small
    \renewcommand{\arraystretch}{1.05}
    \setlength{\tabcolsep}{5pt}
    \caption{\textbf{Main results on Vision2Web and Webdev-Bench.} Vision2Web (L2-frontend, 66 titles) reports the visual score (VS), functional score (FS), their average (Avg), and the delivery success rate (DSR); Webdev-Bench (40 tasks) reports requirement satisfaction (RS) together with its visual (VS), functional (FS), and interaction (IS) components, their average (Avg), and DSR. Undelivered tasks count as $0$ in all score columns. Best in \rankbest{bold} and second best \ranksecond{underlined} per column across all models.}

    \resizebox{\linewidth}{!}{
    \begin{tabular}{lcccccccccc}
    \toprule
    \multirow{2}{*}{\textbf{Model}} & \multicolumn{4}{c}{\textbf{Vision2Web}} & \multicolumn{6}{c}{\textbf{Webdev-Bench}} \\
    \cmidrule(lr){2-5}\cmidrule(lr){6-11}
    & \textbf{VS} & \textbf{FS} & \textit{Avg} & \textit{DSR} & \textbf{RS} & \textbf{VS} & \textbf{FS} & \textbf{IS} & \textit{Avg} & \textit{DSR} \\
    \midrule
    
    \rowcolor{gray!12}
    \multicolumn{11}{c}{
    \textit{\textbf{Proprietary Models}}
    } \\


    Claude-Opus-4.8 &
    \rankbest{59.1} &
    65.9 &
    \rankbest{62.5} &
    98.5\% &
    \ranksecond{65.1} &
    75.0 &
    \ranksecond{70.7} &
    84.1 &
    \ranksecond{73.7} &
    100.0\% 
\\
    GPT-5.4 &
    51.3 &
    64.4 &
    57.9 &
    97.0\% &
    59.6 &
    72.7 &
    70.0 &
    81.0 &
    70.8 &
    100.0\% 
\\
    GPT-5 &
    23.3 &
    64.7 &
    44.0 &
    98.5\% &
    53.5 &
    64.0 &
    65.8 &
    74.6 &
    64.5 &
    100.0\% 
\\
    Gemini-3-Pro &
    13.3 &
    14.6 &
    14.0 &
    65.2\% &
    48.0 &
    65.8 &
    58.5 &
    68.5 &
    60.2 &
    100.0\% 
\\
    Gemini-3-Flash &
    13.1 &
    27.9 &
    20.5 &
    78.8\% &
    56.6 &
    65.8 &
    64.8 &
    \ranksecond{85.0} &
    68.1 &
    95.0\% 
\\
    Seed-1.8 &
    7.0 &
    13.3 &
    10.2 &
    36.4\% &
    52.6 &
    66.5 &
    63.8 &
    71.5 &
    63.6 &
    100.0\% 
\\

    
    \addlinespace[3pt]
    \rowcolor{gray!12}
    \multicolumn{11}{c}{
    \textit{\textbf{Open-Source Models}}
    } \\
    Kimi-K2.5 &
    35.1 &
    61.0 &
    48.1 &
    92.4\% &
    46.8 &
    61.5 &
    55.0 &
    67.7 &
    57.7 &
    82.5\% 
\\

    
    MiMo-V2.5 &
    38.2 &
    \ranksecond{66.5} &
    52.4 &
    87.9\% &
    14.9 &
    25.2 &
    21.2 &
    22.6 &
    21.0 &
    65.0\% 
\\
    Step-3.7-Flash &
    22.1 &
    47.3 &
    34.7 &
    95.5\% &
    10.8 &
    20.0 &
    14.0 &
    19.0 &
    15.9 &
    60.0\% 
\\
    Qwen3.8-27B &
    36.4 &
    45.7 &
    41.1 &
    48.5\% &
    64.8 &
    \ranksecond{77.9} &
    68.8 &
    79.8 &
    72.8 &
    95.0\% 
\\
    Qwen3.5-397B-A17B &
    35.6 &
    \rankbest{67.1} &
    51.4 &
    100.0\% &
    28.7 &
    39.0 &
    31.9 &
    36.5 &
    34.0 &
    80.0\% 
\\
    Qwen3.5-122B-A10B &
    20.2 &
    39.0 &
    29.6 &
    87.9\% &
    24.6 &
    30.4 &
    28.9 &
    33.1 &
    29.3 &
    72.5\% 
\\

    Qwen3-VL-32B-Instruct &
    0.0 &
    0.0 &
    0.0 &
    0.0\% &
    2.5 &
    6.3 &
    5.4 &
    4.4 &
    4.6 &
    50.0\% 
\\
    Qwen3-VL-8B-Instruct &
    0.0 &
    0.0 &
    0.0 &
    0.0\% &
    0.0 &
    0.8 &
    0.0 &
    0.0 &
    0.2 &
    30.0\% 
\\
    \textbf{\ModelName{}-27B} &
    \ranksecond{57.7} &
    \rankbest{67.1} &
    \ranksecond{62.4} &
    80.3\% &
    \rankbest{72.8} &
    \rankbest{81.9} &
    \rankbest{73.8} &
    \rankbest{90.8} &
    \rankbest{79.8} &
    100.0\% 
\\

    \bottomrule
    \end{tabular}}

    \label{tab:main_results}
\end{table}

\subsection{Ablation Studies}
\label{sec:ablation}
We compare the full model with two ablated variants that share its initialization, training data, terminal reward, and hyperparameters, evaluated on Vision2Web L2 with the base model as reference. \emph{\textbf{w/o Compaction (linear truncation)}} treats each rollout as a single linear token sequence and trains only on its first 65{,}536 tokens. \emph{\textbf{w/o Artifact-Grounded Credit Assignment}} keeps the sample-construction procedure but gives all segments of a rollout the same rollout-level GRPO advantage.

\begin{figure}[t]
    \centering
    \includegraphics[width=\linewidth]{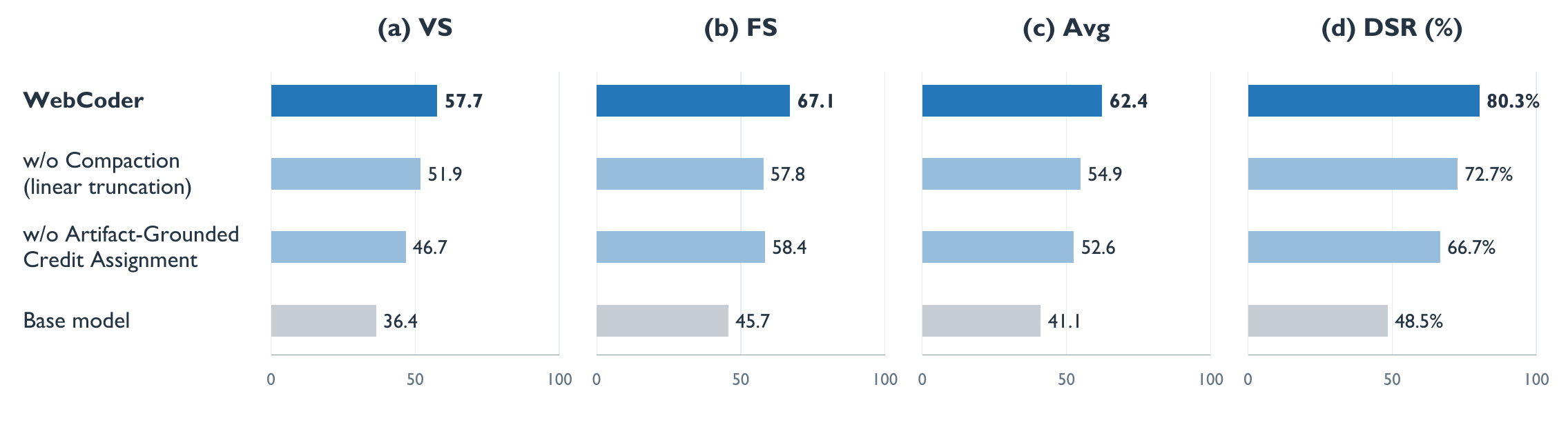}
    \caption{
        \textbf{Ablation studies of \ModelName{} on Vision2Web L2.}
        Full model, the two ablated variants, and the base model on (a) VS, (b) FS, (c) Avg, and (d) DSR. Higher is better.
    }
    \label{fig:ablation-studies}
\end{figure}

Figure~\ref{fig:ablation-studies} shows that the full model outperforms both variants on all four metrics.
Removing compaction reduces Avg from 62.4 to 54.9, a drop of 7.5 points, while removing artifact-grounded credit assignment reduces it to 52.6, a drop of 9.8 points.
Both ablations also lower visual and functional scores: VS and FS decrease by 5.8 and 9.3 points, respectively, without compaction, and by 11.0 and 8.7 points without artifact-grounded credit assignment.
Both variants remain above the base model's Avg of 41.1, but neither matches the full training recipe.
DSR also decreases from 80.3\% to 72.7\% and 66.7\% for the compaction and credit-assignment ablations, respectively.

\subsection{In-depth Analysis}

\begin{figure*}[t]
    \centering
    \includegraphics[width=\textwidth]{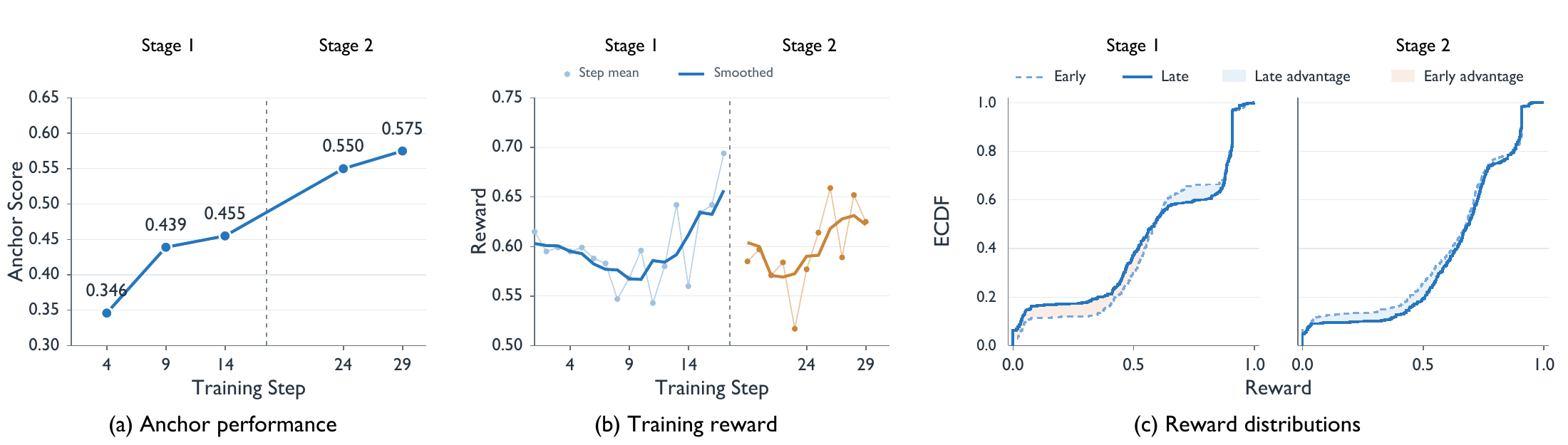}
    \caption{
        \textbf{Training dynamics of \ModelName{}}.
        (a) Evaluation scores on the fixed anchor set.
        (b) Per-step mean training rewards and centered five-step moving averages, computed separately within each stage.
        (c) Early and late reward distributions: steps 1--4 versus 11--14 in Stage 1, and steps 18--21 versus 26--29 in Stage 2. Blue and orange shading mark reward thresholds above which the late or the early window, respectively, has the larger fraction of rollouts.
    }
    \label{fig:training-dynamics}
\end{figure*}

Figure~\ref{fig:training-dynamics} reports evaluation scores on a fixed held-out set of 28 anchor tasks and rewards from on-policy training rollouts; the fixed set gives a common reference across checkpoints, whereas training rewards depend on the tasks sampled at each step. The anchor score increases across all evaluated checkpoints, from 0.346 at step 4 to 0.575 at step 29 (a relative gain of 66.2\%), with improvements within both training stages. Training rewards, in contrast, show an early decline followed by a recovery in Stage 1; behavioral statistics (Appendix~\ref{sec:agent-behavior}) associate the dip on replication tasks with more redundant operations and less frequent skill invocation, both of which reverse during the recovery. Interaction volume alone therefore does not capture development progress, which motivates credit assignment based on changes in the artifact.

Figure~\ref{fig:training-dynamics}(c) shows how reward distributions change within each stage. In Stage 1, the late window (steps 11--14) contains both more high-reward rollouts than the early window (steps 1--4; reward $\geq 0.8$: 33.8\% to 39.7\%) and more zero-reward rollouts (2.6\% to 6.3\%). In Stage 2, the zero-reward fraction decreases from 6.5\% (steps 18--21) to 4.9\% (steps 26--29) while the mean reward increases from 0.604 to 0.633, so the later gains come with fewer failures rather than only more high-reward outcomes.

\subsection{Case Studies}
\label{sec:case-studies}

\begin{figure}[t]
\centering
\includegraphics[width=\linewidth]{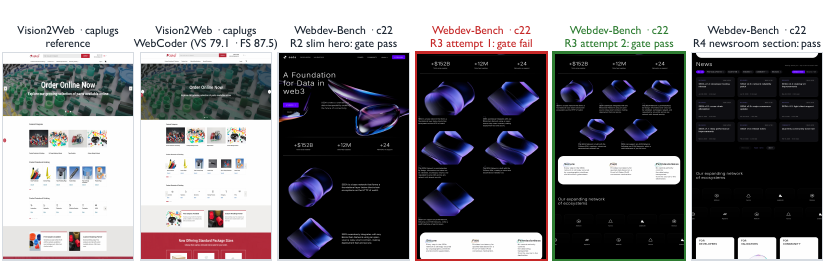}
\caption{\textbf{\ModelName{} on Vision2Web (caplugs) and Webdev-Bench (c22).} Left: the reference homepage of caplugs and \ModelName{}'s render. Right: c22 states after each round, each cropped to the region the requirement changes. R2 tightens the hero and removes the investor row; the first attempt at R3 keeps the five ``Why SEDA'' parts in the old staggered layout and is rejected by the gate, while the second attempt compacts them into two rows and passes; R4 adds a filterable newsroom below the hero.}
\label{fig:case-main}
\end{figure}

Figure~\ref{fig:case-main} shows \ModelName{} on one task from each benchmark. On the Vision2Web title \emph{caplugs}, a six-page e-commerce site, the agent reproduces the reference's hero banner, category grid, and promotional band and serves all six pages (VS $79.1$, FS $87.5$). On the Webdev-Bench task c22, a five-round modification of an existing website, the simulated-user gate rejects the first attempt at requirement three, the agent rearranges the section and passes on the second attempt, and terminal acceptance grades all $27$ checklist items $3$ (RS $1.0$, against $0.6$ for GPT-5.4). Appendix~\ref{app:case-studies} adds baseline comparisons, full trajectories, and a task on which \ModelName{} falls short.

\section{Related Work}

\paragraph{Frontend Development.} Automated frontend development spans reproducing interfaces from visual references, generating applications from textual requirements, and modifying or repairing existing projects. Datasets such as WebSight~\citep{laurenccon2024unlocking}, Web2Code~\citep{yun2024web2code}, and WebCode2M~\citep{gui2025webcode2m} support screenshot-to-code learning, and benchmarks including Design2Code~\citep{si2025design2code}, Sketch2Code~\citep{li2025sketch2code}, Figma2Code~\citep{gui2026figma2code}, and Interaction2Code~\citep{xiao2025interaction2code} evaluate generation from screenshots, sketches, design files, and interactive prototypes. WebGen-Bench~\citep{lu2026webgen-bench} and Vision2Web~\citep{he2026vision2web} add browser-based functional testing, while DesignBench~\citep{xiao2025designbench} and MT-Web2Code~\citep{li2026mt} cover editing and repair with preservation of unaffected content. On the modeling side, UICopilot~\citep{gui2025uicopilot} and ScreenCoder~\citep{jiang2025screencoder} generate code coarse-to-fine, JanusCoder~\citep{sun2026januscoder} unifies generation and editing, UI2Code$^{N}$~\citep{yang2026ui2code} and VisRefiner~\citep{deng2026visrefiner} refine from rendered feedback, and WebWorld~\citep{xiao2026webworld} selects browser-verified transitions for supervised training. Our work instead trains a single policy to compose all four activities within the same evolving application through artifact-grounded agentic reinforcement learning.

\paragraph{Agentic Reinforcement Learning.} Agentic reinforcement learning optimizes policies through multi-step interactions with tools and environments~\citep{wang2025ragen,luo2025agent}; SWE-RL~\citep{wei2026swe} learns from software evolution data with rule-based rewards, and GiGPO~\citep{feng2026group} combines episode-level advantages with step-level estimates from repeated environment states. In frontend development, WebGen-Agent~\citep{lu2026webgen-agent} uses screenshot and GUI-agent scores as step-level rewards, ReLook~\citep{li2026relook} trains refinement with a multimodal critic, and WebGrader~\citep{chen2026webgrader} derives executable interaction checks from requirements. For trajectories beyond the context window, ReSum~\citep{wu2025resum} broadcasts advantages across summarized segments, Context-Folding~\citep{sun2025scaling} learns branching and folding, and CompactionRL~\citep{li2026compactionrl} jointly optimizes execution and summarization. Our approach grounds terminal rewards and intermediate progress estimates in one checklist, redistributes the rollout advantage across segments by changes in these estimates, and trains across context rewrites with message-tree samples and ownership masks.

\section{Conclusion}


We introduce \ModelName{}, a general frontend development agent trained with artifact-grounded agentic RL in Webdev-World. A single policy coordinates reproduction, generation, modification, and repair over a persistent project state; one requirement decomposition grounds both terminal rewards and credit assignment across development stages; and compaction with message-tree samples supports training beyond a single context window. On Webdev-Bench, which we introduce to evaluate capability composition, and on Vision2Web, \ModelName{} improves substantially over its base model and matches or exceeds the strongest proprietary models, supporting artifact evolution under shared requirements as a basis for training general frontend agents.

\section*{Use of Large Language Models}
We used LLM-based coding assistants during this project for code development, data-processing scripts, figure generation, and editing of the manuscript, in the sense described by the ICLR policy. LLMs are also components of the method and the evaluation, and these uses are documented in the paper rather than being incidental: a vision--language judge (GLM-5V-Turbo) forms one leg of the training reward (Section~\ref{sec:method-rl}, Appendix~\ref{app:reward-credit}); an LLM acts as the simulated user and the acceptance judge in the Webdev-Bench protocol (GLM-5.3-Flash, Appendix~\ref{sec:protocol}); LLM-based author, reviewer, and hardening stages assist the construction of Webdev-Bench tasks and hidden checklists, whose outputs were reviewed by the authors (Appendix~\ref{sec:construct}); and the official Vision2Web scorer uses LLM judges (Appendix~\ref{sec:vision2web}). The reliability of the judges is measured in Appendix~\ref{app:verifier} and their agreement with human annotators in Appendix~\ref{app:human-eval}. We did not use LLMs to generate experimental results, fabricate citations, or produce claims without support. The authors reviewed all LLM-assisted text and code and take full responsibility for the content of the paper.

\section*{Ethics Statement}
This work trains and evaluates agents that write frontend code. Training tasks are constructed from synthetic product specifications and from screenshots of publicly accessible web pages used as visual references; Webdev-Bench tasks are likewise seeded by public-page screenshots, synthetic specifications, or repositories prepared by the authors. No user data, account-restricted content, or non-public documents were collected, and reference pages were selected to avoid personal information. Agents run in isolated containers with no access to external services beyond package registries, and the verifiers inspect only the applications the agents build. Models that reproduce web pages from screenshots could be misused to clone protected designs; our benchmark uses reproduction only as the first step of composite development tasks, and the released materials are intended for research on frontend development agents.

\section*{Reproducibility Statement}
The method is specified in Section~\ref{sec:method}: the environment, the terminal reward and its checklist decomposition, the credit-assignment computation, and the policy objective. Appendix~\ref{sec:long-horizon-training} gives the implementation details needed to reproduce training: context management and message-tree sample construction with pseudocode (Algorithm~\ref{alg:message-tree-training}), the evidence, verification method, and weight of each requirement type in the reward, the definition of the intermediate potentials and segment offsets, the rollout infrastructure, and all hyperparameters (Table~\ref{tab:training-config}). Appendix~\ref{app:prompts} reproduces the prompts and skill definitions supplied to the agent, the task-construction prompts, and the acceptance prompt. Appendix~\ref{sec:webdev-bench} describes the benchmarks, the Webdev-Bench construction pipeline, the evaluation protocol with time budgets and retry rules, the judge models, and the exact scoring formulas; Appendix~\ref{app:baselines} lists every baseline and its serving configuration. Every number in the tables and figures is computed programmatically from archived per-task reports and judge outputs by scripts that will be released with the benchmark. Upon acceptance, we will open-source the \ModelName{} model, the Webdev-Bench benchmark, and the training and evaluation code. Judge stochasticity is controlled by median-of-three grading during training and quantified by repeated acceptance passes in Appendix~\ref{app:verifier}.


\bibliography{iclr2027_conference}
\bibliographystyle{iclr2027_conference}


\clearpage
\appendix

\startcontents[appendix]

\section*{Appendix Contents}
\printcontents[appendix]{}{1}{%
    \setcounter{tocdepth}{2}%
}

\clearpage


\section{Baselines}
\label{app:baselines}

We compare \ModelName{}-27B with fourteen models, all evaluated under the same agent framework, tools, prompts, container environment, time budget, and judges as \ModelName{}. 

\paragraph{Proprietary models.}
\textbf{Claude-Opus-4.8}~\citep{anthropic2026opus48} is Anthropic's flagship model. \textbf{GPT-5.4}~\citep{openai2026gpt54} and \textbf{GPT-5}~\citep{openai2025gpt5} are OpenAI's two most recent flagship models. \textbf{Gemini-3-Pro}~\citep{google2025gemini3pro} and \textbf{Gemini-3-Flash}~\citep{google2025gemini3flash} are Google's flagship and efficiency-oriented models of the Gemini 3 family. \textbf{Seed-1.8}~\citep{seed2026seed18} is ByteDance's multimodal model designed for real-world agentic tasks.

\paragraph{Open-source models.}
\textbf{Kimi-K2.5}~\citep{team2026kimi25} is Moonshot AI's visual agentic model. \textbf{MiMo-V2.5}~\citep{xiaomi2026mimo25} is Xiaomi's multimodal model. \textbf{Step-3.7-Flash}~\citep{stepfun2026step37flash} is StepFun's efficiency-oriented model. \textbf{Qwen3.8-27B}~\citep{qwen2026qwen38} is the dense model from which \ModelName{}-27B is initialized and serves as the base-model reference. \textbf{Qwen3.5-397B-A17B} and \textbf{Qwen3.5-122B-A10B}~\citep{qwen2026qwen35} are mixture-of-experts models of the Qwen3.5 family with 17B and 10B activated parameters. \textbf{Qwen3-VL-32B-Instruct} and \textbf{Qwen3-VL-8B-Instruct}~\citep{yang2025qwen3} are vision--language instruction models of the Qwen3 family.

\section{Benchmarks}

\subsection{Vision2Web}
\label{sec:vision2web}
Vision2Web~\citep{he2026vision2web} is a benchmark for evaluating multimodal coding agents on website development, covering static webpages, interactive frontends, and full-stack websites.
We evaluate on the Interactive Frontend level (L2), which contains 66 tasks.
Each task provides multiple UI prototypes, textual requirements, and supporting assets, and requires an interactive frontend with visual consistency and coherent navigation across pages.
This level assesses both visual reproduction and functional implementation, aligning with our focus on frontend development.

We allow two hours of development per task and independently deploy and evaluate the resulting projects.
During deployment, the project's startup script is executed in a fresh container, and the application is checked for accessibility at the designated address.
Once deployment succeeds, evaluators interact with the application according to predefined workflows, checking functional outcomes and collecting page screenshots.
Functional evaluation uses Claude Sonnet 4.6 with \texttt{playwright-cli} to execute interactions and record test-case outcomes.
Visual evaluation uses GLM-4.6V\textcolor{blue}{~\citep{zai2026glm46v}} to compare the captured screenshots against the reference prototypes.
The same evaluation configuration is used for all models.

We report Visual Score (VS), Functional Score (FS), their average (Avg), and Deployment Success Rate (DSR).
VS aggregates visual scores across reference prototypes, while FS measures the proportion of functional test cases passed; both are reported on a 0--100 scale.
Avg is the arithmetic mean of VS and FS, and DSR is the proportion of the 66 tasks whose applications are successfully deployed.
All tasks are included in the reported results, with failed deployments contributing zero scores for their associated prototypes and functional test cases.
These metrics jointly characterize deployability, visual fidelity, and functional correctness.

\subsection{Webdev-Bench}
\label{sec:webdev-bench}

\begin{figure}[htbp]
\centering
\includegraphics[width=\linewidth]{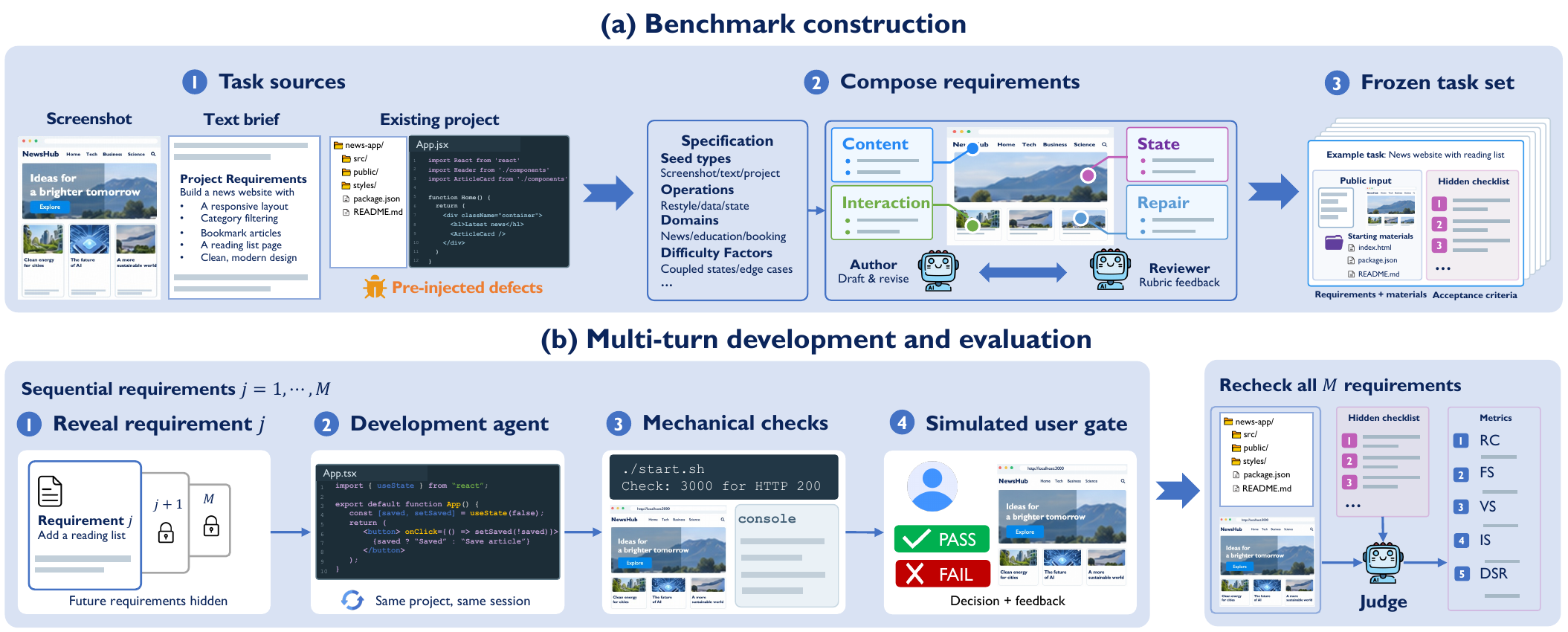}
\caption{\textbf{Construction (top) and multi-round evaluation (bottom) of Webdev-Bench.} During construction, a specification drives a two-agent authoring loop that produces each task together with its hidden checklist. During evaluation, requirements are revealed round by round under a simulated-user gate, and an independent acceptance agent scores the complete application at the end.}
\label{fig:webdev-bench}
\end{figure}

Webdev-Bench evaluates an agent's ability to develop a real front-end project continuously. The benchmark comprises 40 tasks spanning three starting conditions: replicating a page from a reference screenshot, building an application from a textual product description, and continuing development on an existing repository, including defect repair. Every task is a composite task: four to five natural-language development requirements are revealed one round at a time, and the agent completes them sequentially within a single session on a single project. Each requirement exercises one concrete development operation (replication, data wiring, interaction, layout restructuring, re-theming, and so on), and together they form one sustained, multi-requirement delivery. After all requirements have been revealed and development ends, the final application undergoes a single terminal acceptance pass that re-examines not only the last requirement but the behavior promised by every earlier one. The central question the benchmark asks is therefore explicit: \emph{can the agent satisfy each newly revealed requirement while preserving the behavior the project has already established?} Figure~\ref{fig:webdev-bench} summarizes the construction and evaluation pipeline; the following subsections cover task composition (\S\ref{sec:stats}), construction (\S\ref{sec:construct}), the evaluation protocol and metrics (\S\ref{sec:protocol}), and two representative tasks (\S\ref{sec:demos}).

\subsubsection{Task Composition and Checklist Statistics}
\label{sec:stats}

\begin{figure}[htbp]
\centering
\includegraphics[width=\linewidth]{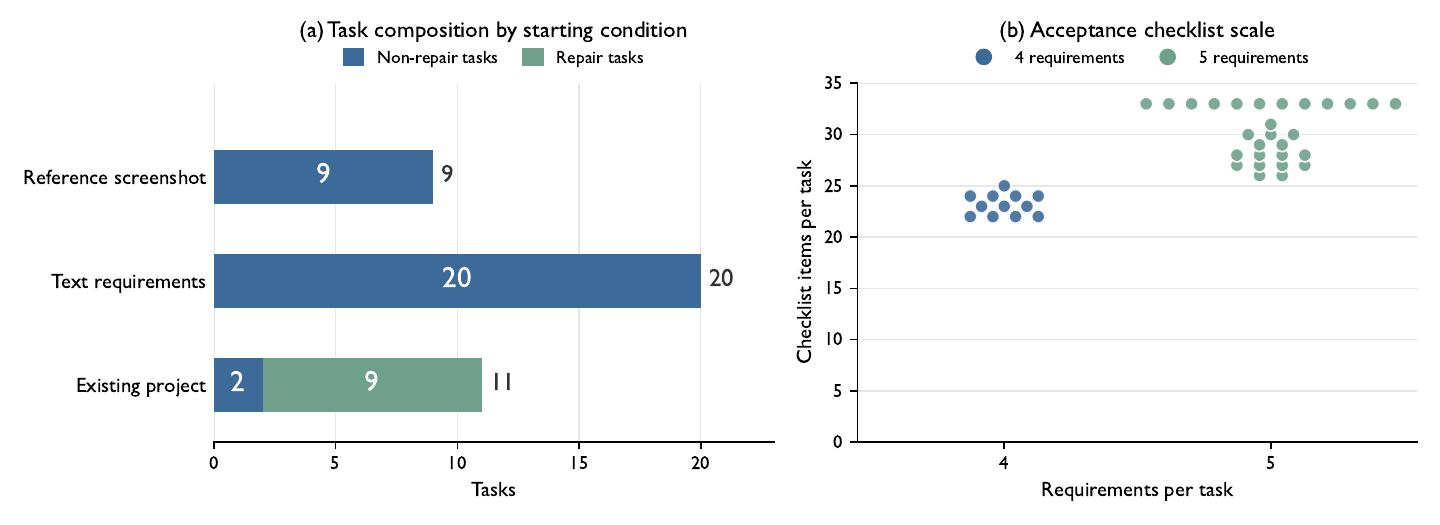}
\caption{\textbf{Task composition and acceptance scale of Webdev-Bench.} (a) Tasks per starting condition, 40 composite tasks in total; the nine repair tasks (green segment) are contained within the existing-project category. (b) Each point is one task, with the number of requirements on the horizontal axis and the number of checklist items on the vertical axis; small horizontal jitter reduces overlap. The benchmark contains 1{,}125 checklist items in total.}
\label{fig:stats}
\end{figure}

All 40 tasks are composite and fall into three starting conditions (Figure~\ref{fig:stats}a). Nine reference-screenshot tasks ask the agent to rebuild a page from a full-page screenshot of a real website and then keep developing on top of it; twenty text-requirement tasks build an application from scratch against a product description; eleven existing-project tasks start from a running repository, and nine of these contain a pre-injected, reproducible defect whose diagnosis and repair constitute the first requirement. The 40 tasks contain 188 requirements in total (five in each of 28 tasks, four in each of the remaining 12). The requirement vocabulary covers eleven operation types, ranging from visually oriented replication and re-theming, through functionally oriented data wiring, stateful behavior, and interaction, to structurally oriented layout restructuring and section addition or removal; task subjects span ten domains including e-commerce, booking services, editorial content, and data dashboards.

During construction, each requirement is further decomposed into checklist items: observable, operationally verifiable descriptions of application behavior, each annotated with its parent requirement, an evaluation dimension (functional, interaction, visual, or coverage), a weight (1.5 for core items, 1.0 otherwise), and an acceptance criterion. The items are hidden from the agent under evaluation throughout; we refer to them collectively as the hidden checklist. Requirement counts and item counts measure scale at two different levels: one requirement typically constrains page content, functional results, and interaction states simultaneously, unfolding into roughly six items on average. Each task carries 22--33 items, and the full benchmark carries 1{,}125 (Figure~\ref{fig:stats}b): 408 functional, 309 visual, 299 interaction, and 108 coverage items (plus one data-consistency item). Coverage items explicitly test whether behavior delivered for earlier requirements still holds after later modifications, realizing the ``preserve established behavior'' objective at the acceptance level. In sum, the task level characterizes sustained development through progressively revealed requirements, while the acceptance level concretizes those requirements into individually checkable application behaviors.

\subsubsection{Benchmark Construction}
\label{sec:construct}

Tasks and their hidden checklists are produced jointly by one pipeline, so that requirement text and acceptance criteria correspond item by item. Construction starts from a specification that fixes a task's starting condition, operation sequence, and difficulty tier (15 regular, 25 challenge), controlling the composition of the exam at the source. Seed material is then drawn from a curated pool according to the specification: replication tasks receive a full-page screenshot of a real website, and existing-project tasks receive a runnable repository; text-requirement tasks need no external seed. The specification and seed are given to an \textbf{author agent}, which drafts the round-by-round natural-language requirements together with the matching hidden checklist; a \textbf{reviewer agent} reads the draft independently and returns revision suggestions concerning feasibility, ambiguity, and acceptance blind spots, which the author incorporates. The purpose of the two-agent loop is to couple ``how a user would phrase the next request'' with ``what acceptance should later check.''

\subsubsection{Evaluation Protocol and Metrics}
\label{sec:protocol}

The protocol deliberately separates two roles: driving development forward and producing the final score. Development proceeds in rounds. When requirement $j$ is revealed, the agent continues working in the same session on the same project, with all later requirements still unknown. At the end of a round, the harness first performs a mechanical health check---launching the application with the task's own script, probing the service port, and collecting a full-page screenshot, the page text, and console output. A simulated user (GLM-5.3-Flash~\citep{zai2026glm53flash}, an LLM independent of the agent) then judges, from this evidence alone, whether the checklist items of the current requirement appear satisfied and issues a pass-or-return verdict with a short critique in the user's voice that must not quote the items: a pass reveals the next requirement, while a return sends the agent back to rework the current one, with at most three attempts per requirement. All rounds share a time budget proportional to the number of requirements (1{,}200 seconds per requirement), with gating and feedback overhead charged against it. 
This gate exists to push development forward in a manner close to a real user; it never sees the hidden checklist and produces no score.

All scores come from terminal acceptance. After development ends, an acceptance agent (also GLM-5.3-Flash) independent of the simulated user operates the delivered application in a browser and grades every item of every requirement against the hidden checklist on a 0--3 scale; requirements already passed by the gate during development are re-examined all the same. Aggregation follows a per-requirement conjunctive rule: a requirement counts as satisfied only if all items under it reach the top grade, and the per-task satisfaction fraction, macro-averaged over the 40 tasks, is the requirement satisfaction score (RS). Restricting the same grade records to the items of each dimension yields the functional (FS), interaction (IS), and visual (VS) scores; the delivery success rate (DSR) is the fraction of tasks that deliver a runnable application within budget, with all items of undelivered tasks graded 0. The two halves of the protocol thus map exactly onto the objective stated in \S\ref{sec:webdev-bench}: round-by-round feedback drives new requirements toward completion, while the terminal full re-examination checks whether later modifications broke earlier results.

Formally, let $R_t$ denote the requirement set of task $t$ and $I_r$ the checklist-item set of requirement $r$; each item $i$ carries a dimension $d_i \in \{\text{functional}, \text{interaction}, \text{visual}, \text{coverage}\}$ and a weight $w_i \in \{1.0, 1.5\}$, and terminal acceptance assigns it a grade $g_i \in \{0,1,2,3\}$, with 3 the top grade. Under the per-requirement conjunctive rule, a requirement is satisfied if and only if every item under it reaches the top grade:
\begin{equation}
\mathrm{sat}(r) \;=\; \mathbf{1}\!\left[\, g_i = 3,\;\ \forall\, i \in I_r \,\right],
\label{eq:sat}
\end{equation}
and RS macro-averages requirement satisfaction within each task and then across the 40 tasks:
\begin{equation}
\mathrm{RS} \;=\; \frac{1}{40} \sum_{t} \frac{1}{|R_t|} \sum_{r \in R_t} \mathrm{sat}(r).
\label{eq:rs}
\end{equation}
Dimension scores restrict each requirement's item set to a dimension subset $D$ and apply the same conjunctive rule: with $I_r^{D} = \{\, i \in I_r : d_i \in D \,\}$ and $\mathrm{sat}_D(r) = \mathbf{1}[\, g_i = 3,\ \forall\, i \in I_r^{D} \,]$, counting only requirements with $I_r^{D} \neq \varnothing$,
\begin{equation}
\mathrm{Score}_D \;=\; \frac{1}{40} \sum_{t} \frac{1}{|\{ r \in R_t : I_r^{D} \neq \varnothing \}|} \sum_{\substack{r \in R_t \\ I_r^{D} \neq \varnothing}} \mathrm{sat}_D(r),
\label{eq:dim}
\end{equation}
where FS takes $D = \{\text{functional}, \text{coverage}\}$---coverage items test whether behavior delivered for earlier requirements still holds after later changes and therefore count toward functional correctness---while IS takes $D = \{\text{interaction}\}$ and VS takes $D = \{\text{visual}\}$. DSR is the number of delivered tasks divided by 40, where a task counts as delivered if its own start script brings the application up on the designated port and passes the launch check; all items of undelivered tasks enter with $g_i = 0$, so their task-level RS and dimension scores are all 0. As a reference scale, the weighted four-grade score of requirement $r$ on subset $D$ is $\sum_{i \in I_r^{D}} w_i \, (g_i / 3) \big/ \sum_{i \in I_r^{D}} w_i$, aggregated exactly as in Eq.~\eqref{eq:dim}.

\subsubsection{Representative Tasks}
\label{sec:demos}

Two tasks from the official set serve as case studies of task design: each is presented through its own initial material, the round-by-round requirement texts (abridged), and an excerpt of its hidden checklist, showing how requirements depend on one another and how the checklist concretizes natural-language requests into state-level checks. Requirement and checklist texts in the figures are abridged and translated from the Chinese originals.

\begin{figure}[htbp]
\centering
\includegraphics[width=\linewidth]{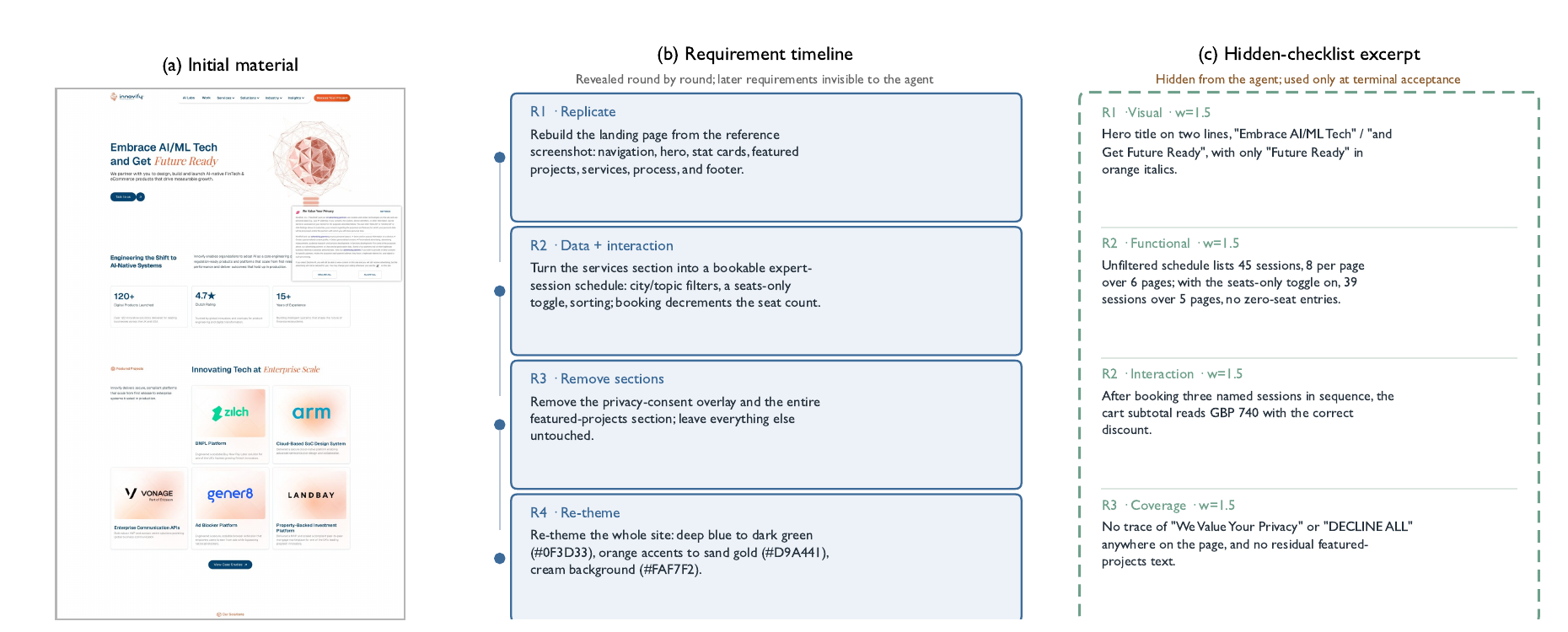}
\caption{\textbf{Anatomy of representative task c38, a screenshot-seeded composite task.} (a) Initial material: the task's own full-page reference screenshot of a real website. (b) The four requirements revealed round by round (abridged; later requirements are invisible to the agent when requirement $j$ is revealed). (c) Hidden-checklist excerpt with parent requirement, dimension, and weight tags; hidden from the agent throughout and used only at terminal acceptance. Texts are translated from the Chinese originals.}
\label{fig:demo38}
\end{figure}

\paragraph{A screenshot-seeded composite task (c38).}
The only initial material is a full-page screenshot of a consulting-firm website, and four requirements are revealed in order (Figure~\ref{fig:demo38}a--b): R1 replicates the landing page from the screenshot; R2 converts the static services section into a bookable expert-session schedule with city and topic filters, a seats-only toggle, sorting, and seat counts that decrement on booking; R3 removes the first-screen privacy-consent overlay and the featured-projects section; R4 re-themes the whole site from deep blue to dark green with warm sand-gold accents on a cream background. The requirements depend on one another: the R2 schedule is built on the layout replicated in R1, and the R4 re-theme must also cover the interactive section newly added by R2. The matching hidden checklist goes beyond ``the section exists'' (Figure~\ref{fig:demo38}c): a functional item under R2 requires that ``with no filters the schedule lists 45 sessions, 8 per page over 6 pages, dropping to 39 sessions over 5 pages when the seats-only toggle is on,'' and an interaction item requires that ``booking three named sessions in sequence yields a subtotal of \pounds 740 with the correct discount applied''---both verifying post-interaction application state rather than the mere presence of controls. A coverage item under R3 requires that no title, body text, or button of the overlay survives anywhere on the page, turning ``remove'' into a criterion that can be checked verbatim.

\begin{figure}[htbp]
\centering
\includegraphics[width=\linewidth]{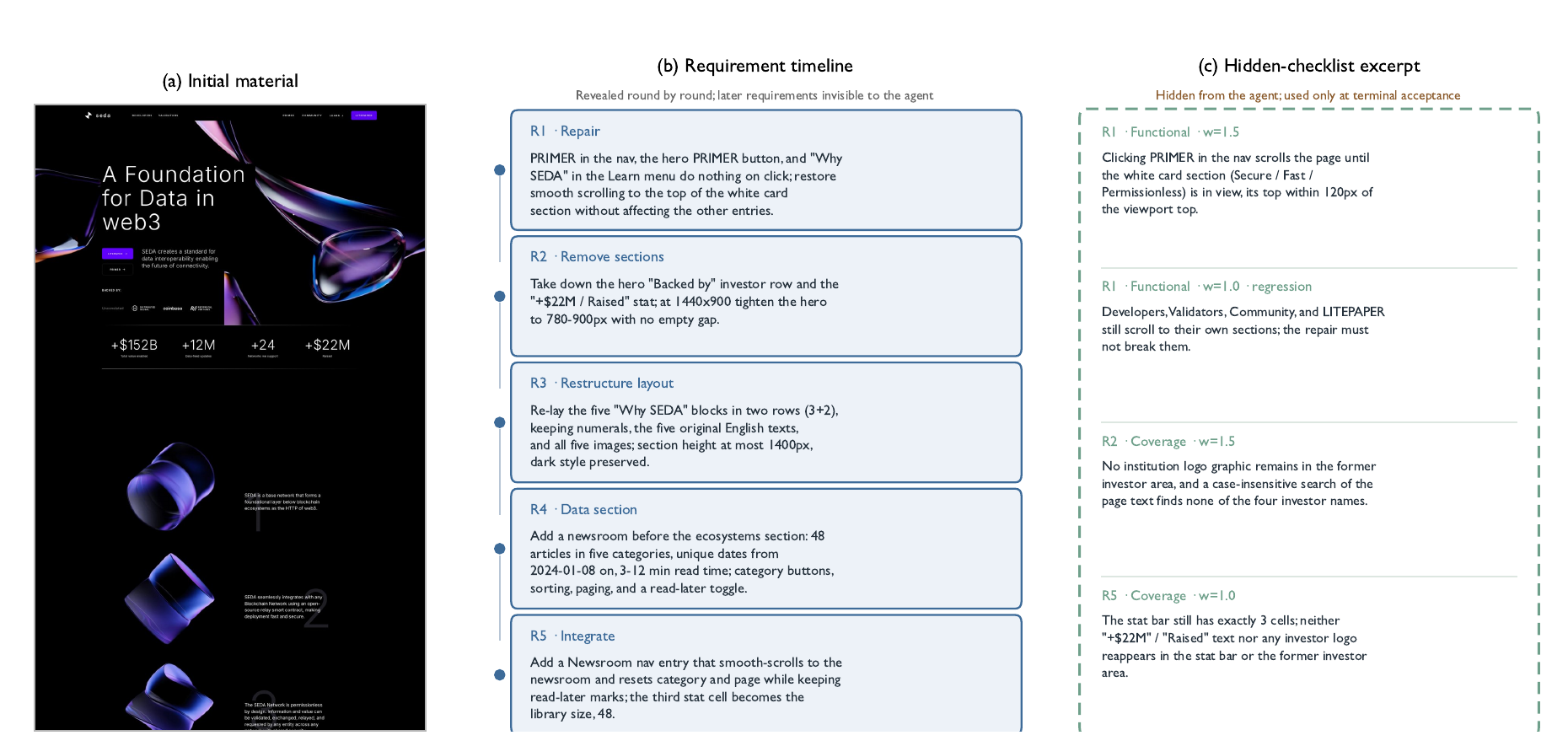}
\caption{\textbf{Anatomy of representative task c22, an existing-project task with defect repair.} (a) Initial material: the above-the-fold view of the task's own starting repository (the injected selector-drift defect disables the three PRIMER entries and is invisible in a static screenshot). (b) The five requirements revealed round by round (abridged). (c) Hidden-checklist excerpt, including the repair item, an explicitly regression-tagged item, and coverage items that check that the removed content does not resurface after later modifications. Texts are translated from the Chinese originals.}
\label{fig:demo22}
\end{figure}

\paragraph{An existing-project task with defect repair (c22).}
The task starts from a runnable website of a web3 data protocol (Figure~\ref{fig:demo22}a, the starting repository's home page) whose repository carries a pre-injected \emph{selector drift} defect: the anchor identifier of the white card section has been altered, so the PRIMER link in the navigation bar, the PRIMER button in the hero, and ``Why SEDA'' in the Learn menu no longer respond to clicks, while the neighbouring entries still scroll smoothly. Five requirements follow: R1 repairs the three entries and confirms that the others are unaffected; R2 removes the hero's investor row and the funding stat and tightens the hero to 780--900 pixels; R3 re-lays the five ``Why SEDA'' blocks in two rows while keeping numerals, original texts, and images; R4 adds a 48-article newsroom before the ecosystems section, with categories, sorting, paging, and a read-later toggle; R5 wires a Newsroom entry into the navigation and turns the third stat cell into the library size (Figure~\ref{fig:demo22}b). The requirements are layered: the R5 entry reuses the scrolling mechanism repaired in R1, its number must agree with the actual article count from R4, and the content removed in R2 must not resurface in R5. The hidden checklist turns these into state-level checks (Figure~\ref{fig:demo22}c): a functional item under R1 requires that ``after clicking PRIMER the card section is in view with its top within 120 pixels of the viewport top,'' accompanied by an explicitly regression-tagged item requiring that the other four entries keep working; a coverage item under R2 requires that no institution logo is rendered in the former investor area and that no investor name appears anywhere in the page text; and a coverage item under R5 searches again for ``+\$22M / Raised'' and investor logos, verifying that the removal still holds after three further rounds of modification rather than merely that it happened.

\section{Long-Horizon Training Implementation}
\label{sec:long-horizon-training}


\subsection{Context Management and Sample Construction}
\label{sec:context-samples}
This section describes how we support development trajectories beyond the context window and construct training samples from them. 
During rollout, context management allows the agent to continue development within a bounded window. 
For training, we select samples from the recorded conversation histories, handle their length limits, and determine which generated tokens contribute to the loss in each sample. 
We describe these steps below and summarize the procedure in Algorithm~\ref{alg:message-tree-training}.

\begin{figure}[h]
\centering
\includegraphics[width=\linewidth]{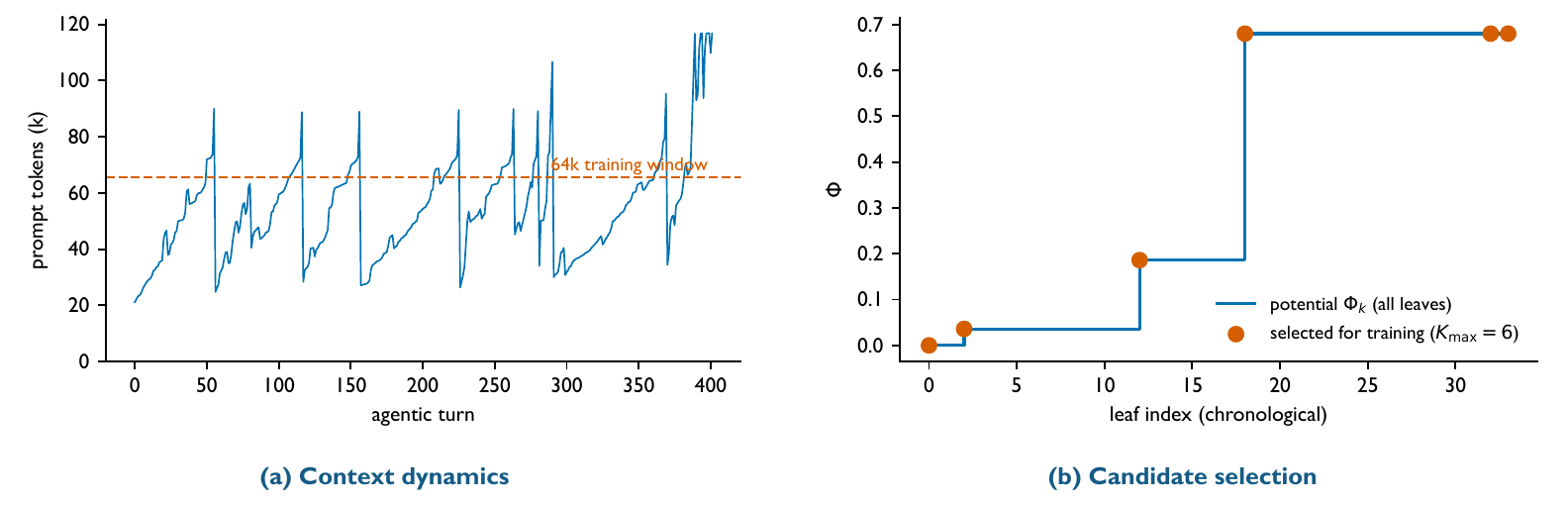}
\caption{\textbf{Context management and training sample selection.}
(a) Prompt length across interaction turns in an example rollout. Context reduction allows development to continue as the history grows; the dashed line marks the 64k training window.
(b) Artifact potential $\Phi$ over chronologically ordered candidate paths. Orange markers indicate the six candidates selected for training from 34 paths, retaining the initial and terminal candidates along with selected intermediate candidates.}
\label{fig:context-management-selection}
\end{figure}

As tool calls accumulate, the conversation history grows. 
We control its length by applying individual tool-output truncation, historical tool-result folding, and summary compaction in sequence. 
After compaction, the agent continues from the shortened history while the project files remain in the environment.
Figure~\ref{fig:context-management-selection}a illustrates this process in one rollout: prompt length increases as interactions accumulate and decreases when the context is reduced. 

\begin{wrapfigure}{r}{0.5\linewidth}
    \centering
    \includegraphics[width=\linewidth]{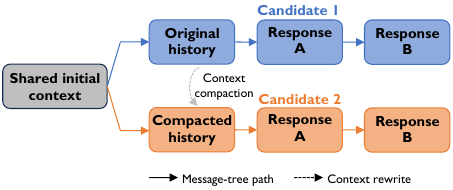}
    \caption{
        \textbf{Message-tree representation}.
        Blue and orange paths show candidates before and after
        compaction. The dashed arrow denotes the context rewrite.
    }
    \label{fig:message-tree}
\end{wrapfigure}

We record the conversation in a message tree to preserve histories before and after context rewrites. 
Shared messages form a common prefix, while rewritten contexts form separate branches.
As illustrated in Figure~\ref{fig:message-tree}, the history before
compaction remains available, and subsequent responses extend the
branch containing the compacted context. Each root-to-leaf path
provides a candidate training sequence containing the responses
along that path and their context.

\Needspace*{16\baselineskip}

To limit training cost, we retain at most $K_{\max}$ candidate paths from each rollout. 
We always retain the initial and terminal candidates, then select the remaining candidates according to measured increases in artifact potential.
Figure~\ref{fig:context-management-selection}b shows an example in which six of 34 candidate paths are retained. 
The curve shows the recorded artifact potential, and the orange markers indicate the selected candidates.

Only eligible model-generated tokens contribute to the training loss; tool responses and other context tokens are excluded.
Since different selected paths may contain the same generated history, we identify repeated generated tokens using message position and content and assign each token to one sample for loss computation.
Its occurrences in other samples serve only as context.
The final loss mask is $M_{k,u}=e_{k,u}\,\mathbf{1}[O(q_{k,u})=k]$, where $e_{k,u}$ indicates whether token $u$ in sample $k$ is eligible for training, $q_{k,u}$ identifies the original generated token, and $O$ specifies its owning sample.

\begin{algorithm}[t]
\caption{Message-Tree Sample Construction for Long-Horizon Training}
\label{alg:message-tree-training}
\begin{algorithmic}[1]
\Require A group of $N$ rollouts for task $x$,
         $\{(\mathcal{T}^{(i)}, R^{(i)}, \mathcal{P}^{(i)})\}_{i=1}^{N}$;
         sample cap $K_{\max}\geq 2$; length limit $L_{\max}$
\Ensure Training batch $\mathcal{B}$

\State $\mathcal{B}\gets\varnothing$
\State $\mu_x\gets \operatorname{Mean}(\{R^{(i)}\}_{i=1}^{N})$
\State $\sigma_x\gets \operatorname{Std}(\{R^{(i)}\}_{i=1}^{N})$

\For{$i=1,\ldots,N$}
    \State $\widehat A^{(i)}
        \gets (R^{(i)}-\mu_x)/(\sigma_x+\epsilon)$
    \State $\mathcal{L}\gets
        \Call{ChronologicalLeafPaths}{\mathcal{T}^{(i)}}$
    \State $K\gets\min(K_{\max},|\mathcal{L}|)$
    \State $\mathcal{S}\gets
        \{\operatorname{First}(\mathcal{L}),
          \operatorname{Last}(\mathcal{L})\}$

    \If{\Call{SelectionScoresAvailable}{$\mathcal{P}^{(i)}$}}
        \State $\mathcal{S}\gets\mathcal{S}\cup
            \Call{TopProgress}{
                \mathcal{L}\setminus\mathcal{S},
                K-|\mathcal{S}|,\mathcal{P}^{(i)}}$
    \Else
        \State $\mathcal{S}\gets\mathcal{S}\cup
            \Call{EvenlySpaced}{
                \mathcal{L}\setminus\mathcal{S},
                K-|\mathcal{S}|}$
    \EndIf

    \For{each path $k\in\mathcal{S}$}
        \State $(z_k,e_k)\gets
            \Call{Assemble}{k,L_{\max}}$
        \Comment{Tokens and eligibility mask}
    \EndFor

    \State $\mathcal{S}\gets
        \{k\in\mathcal{S}:\sum_u e_{k,u}>0\}$
    \State $O\gets
        \Call{AssignOwnership}{
            \mathcal{S},\{z_k,e_k\}_{k\in\mathcal{S}}}$

    \For{each $k\in\mathcal{S}$}
        \State $M_{k,u}\gets
            e_{k,u}\,
            \mathbb{1}[O(q_{k,u})=k]$
        \State $n_k\gets\sum_u M_{k,u}$
    \EndFor
    \State $\mathcal{S}\gets\{k\in\mathcal{S}:n_k>0\}$

    \If{$\mathcal{S}=\varnothing$}
        \State \textbf{continue}
    \EndIf

    \State $\{\delta_k\}_{k\in\mathcal{S}}\gets
        \Call{CreditOffsets}{
            \mathcal{P}^{(i)},\mathcal{S},\{n_k\}_{k\in\mathcal{S}}}$
    \Comment{$\sum_k n_k\delta_k=0$; Appendix~\ref{app:reward-credit}}

    \For{each $k\in\mathcal{S}$}
        \State $\widehat A_k\gets\widehat A^{(i)}+\delta_k$
        \State $\mathcal{B}\gets\mathcal{B}\cup
            \{(i,z_k,M_k,\widehat A_k)\}$
    \EndFor
\EndFor

\State \Return $\mathcal{B}$
\end{algorithmic}
\end{algorithm}

\subsection{Reward Modeling and Credit Assignment Details}
\label{app:reward-credit}
This section specifies the verifiers used during training and the credit-assignment computation, complementing Section~\ref{sec:method-rl} and Algorithm~\ref{alg:message-tree-training}.

\paragraph{Terminal reward.} The verifier builds the project, resolves its entry point, and renders it in a browser; the product of these three gates is $G\in\{0,1\}$. The quality term combines a deterministic leg and a vision--language judge leg (GLM-5V-Turbo~\citep{hong2026glm}), $R=G\cdot(0.6\,f_{\mathrm{det}}+0.4\,f_{\mathrm{vlm}})$, with a penalty of $0.1$ when the wall-clock budget is exhausted. Table~\ref{tab:reward-evidence} lists, for each requirement type, the evidence the verifier collects, how it is verified, and its weight. The judge grades a rendered screenshot on a five-level anchored scale per dimension (mapped to $1.0/0.75/0.5/0.25/0$), and its score is the median of three draws.

\begin{table}[t]
\centering
\small
\caption{\textbf{Evidence, verification method, and weight of each requirement type in the training reward.} Weights of judge dimensions are given as the judge-leg weight $0.4$ times the within-leg dimension weight.}
\label{tab:reward-evidence}
\resizebox{\linewidth}{!}{\begin{tabular}{llp{5.2cm}c}
\toprule
Requirement type & Evidence & Verification & Weight \\
\midrule
\multicolumn{4}{l}{\emph{Reproduction tasks (reference screenshot)}} \\
Pixel-level fidelity & rendered page vs.\ reference & deterministic image comparison ($f_{\mathrm{det}}$) & 0.6 \\
Visual fidelity / layout & screenshot & judge, anchored scale & $0.4\times0.30$ / $0.22$ \\
Content coverage / text match & screenshot & judge, anchored scale & $0.4\times0.24$ / $0.10$ \\
Render health & screenshot & judge, anchored scale & $0.4\times0.14$ \\
\midrule
\multicolumn{4}{l}{\emph{Generation tasks (textual requirements)}} \\
Required sections and counts & DOM census of the served page & keyword presence and graded counts per required section ($f_{\mathrm{det}}$) & 0.6 \\
Design quality / content quality & screenshot & judge, anchored scale & $0.4\times0.38$ / $0.26$ \\
Functionality (visual cues) & screenshot & judge, anchored scale; interactions are not executed & $0.4\times0.18$ \\
Render health & screenshot & judge, anchored scale & $0.4\times0.18$ \\
\midrule
Build, entry point, render & build log, served port, browser & hard gates, multiplicative & $G$ \\
\bottomrule
\end{tabular}}
\end{table}

\paragraph{Intermediate potentials.} Two instantiations of $\phi_{k,i}$ are used. For tasks with a requirement checklist, $\phi_{k,i}$ is a ledger value: whenever the agent fetches a page from its own running application, the fetched text is scored against requirement $i$'s expected content with the same matching function as the terminal $f_{\mathrm{det}}$, and the ledger takes the new value; $\phi_{k,i}$ is the ledger state at the last fetch at or before boundary $k$. Missing measurements carry the previous value forward, and a lower measurement is recorded in full. File reads and writes are not observations, and the checklist is hidden from the agent. Count-based checks require the terminal DOM census and are excluded from the ledger. For reproduction tasks, which have no checklist, a source-authoring score over the files present at boundary $k$ is used: whether the application root is no longer the scaffold placeholder ($0.30$), authored source lines ($0.20$), component files ($0.20$), local modules actually imported ($0.15$), and styling ($0.15$), each saturating; a file is attributed to the earliest segment whose messages name it.

\paragraph{Segment offsets.} For a rollout with segments $k=0,\dots,K{-}1$ and potentials $\Phi_k$, the shaped interior rewards are $r_k=\gamma\Phi_{k+1}-\Phi_k$ for $k<K{-}1$ and $r_{K-1}=0$; the offsets are the discounted returns $b_k=\sum_{j\ge k}\gamma^{j-k}r_j$ with $\gamma=0.995$, centered to a token-weighted mean of zero, clipped to $\pm5$, and re-centered. The rollout's total advantage contribution under the token-weighted loss therefore equals $\widehat A_{\mathrm{rollout}}$, and potentials only redistribute credit within a rollout. If any segment lacks a potential, the rollout falls back to equal broadcast.

\paragraph{Segment selection and ownership.} Each rollout keeps at most $K_{\max}=6$ segments for training: the first, which trains the shared trunk; the last, which holds the terminal artifact; and the interior segments with the largest $\Delta\Phi_k$, falling back to evenly spaced segments when potentials are unavailable. Ownership is keyed by message position and content: a generated turn is trained by the first retained path that contains it and appears as loss-masked context in later paths, so every retained token is trained exactly once with the advantage of the segment in which it was generated. Tokens of a dropped segment are trained by the next retained path and take that path's advantage. Rewards are split as $R/K$ across the $K$ retained samples and re-summed per rollout before group statistics, so each rollout is one data point in GRPO.

\paragraph{Training window and log-probabilities.} Rollouts use a $116{,}736$-token window and training samples are capped at $65{,}536$ tokens. A path longer than the cap keeps its most recent $65{,}536$ tokens, so the trainable segment is preserved and the earliest history is dropped. Behavior-policy log-probabilities are recomputed by the training engine on the assembled sequence; rollout-engine log-probabilities are recorded to monitor the discrepancy between the two engines. Other settings follow Table~\ref{tab:training-config}, with the clipping threshold $\eta=0.2$.

\subsection{Rollout Infrastructure}
\label{sec:rollout-infra}
Agent containers execute rollouts concurrently, while rollout collection and policy optimization alternate on shared GPU devices.
Updated policy weights are synchronized before the next rollout phase.
We use speculative decoding with the model's native MTP head and multi-turn prefix caching to improve serving efficiency~\citep{leviathan2023fast,zheng2024sglang}.

We monitor the log-probability differences between the rollout and training engines on the same generated tokens and support truncated importance sampling (TIS) as an optional correction~\citep{yao2025rollout}.
Transient trajectory-decoding failures are retried with backoff.
Rollouts invalidated by infrastructure failures are excluded from training, while valid task failures are scored by the terminal verifier.

\subsection{Training Configuration}
\label{sec:training-config}

We initialize the policy from Qwen3.8-27B and train it with GRPO.
Each training step samples 16 tasks with eight rollouts per task.
We use a constant learning rate of $2\times10^{-6}$ and BF16
precision. 
Evaluation is performed every five training steps on a fixed set of 28 tasks.
Table~\ref{tab:training-config} summarizes these settings and the sequence length limits.

\begin{table}[t]
    \centering
    \small
    \renewcommand{\arraystretch}{1.15}
    \caption{Model initialization and hyperparameter settings for long-horizon RL training.}
    \begin{tabular}{@{}ll@{}}
        \toprule
        \textbf{Item} & \textbf{Configuration} \\
        \midrule
        Policy initialization & Qwen3.8-27B \\
        Training algorithm & GRPO \\
        Tasks per training step & 16 \\
        Rollouts per task & 8 \\
        Learning rate & $2\times10^{-6}$ (constant) \\
        Training precision & BF16 \\
        Training context limit & 65,536 tokens \\
        Rollout context limit & 116,736 tokens \\
        Maximum output per turn & 16,000 tokens \\
        Evaluation interval & Every 5 training steps \\
        Evaluation set & 28 fixed tasks \\
        \bottomrule
    \end{tabular}
    \label{tab:training-config}
\end{table}

\section{Trajectory Scale and Rollout Efficiency}
\begin{figure}[t]
\centering
\includegraphics[width=\linewidth]{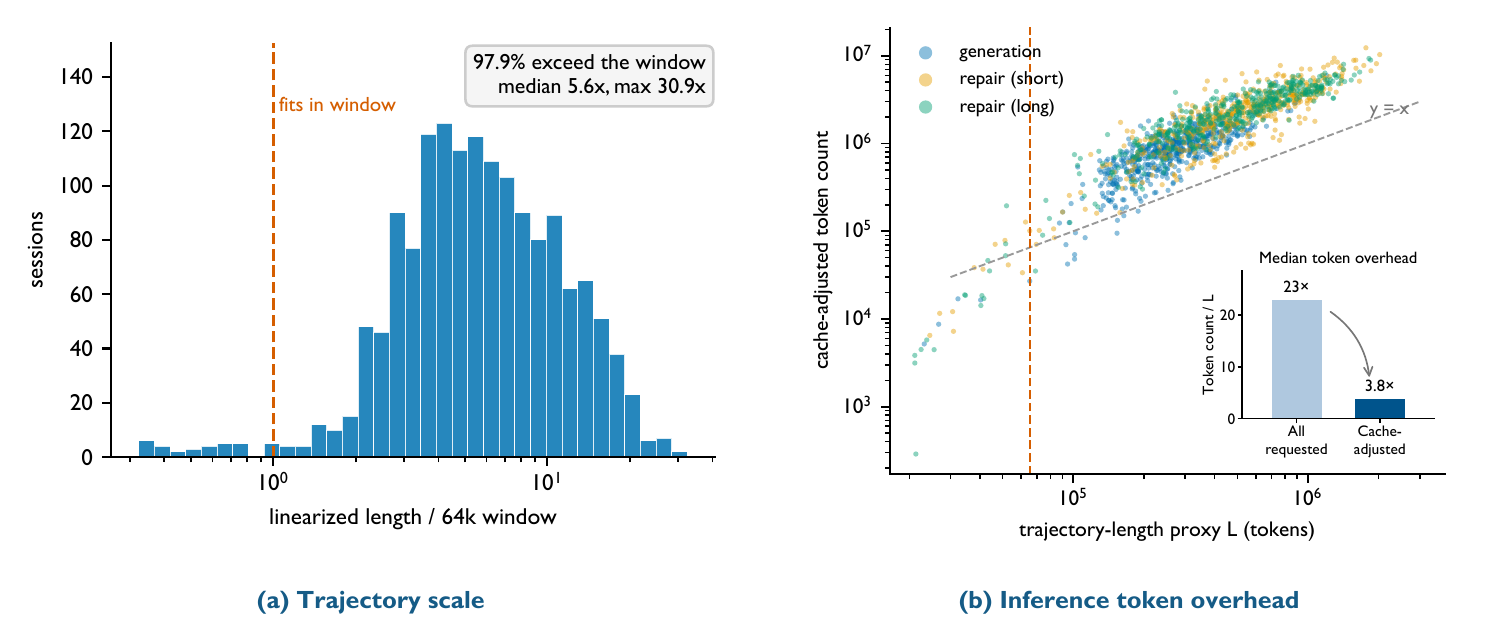}
\caption{\textbf{Trajectory scale and inference token overhead.}
(a) Distribution of the trajectory-length proxy normalized by the 64k training window across 1,556 rollout sessions. The dashed line marks one window; 97.9\% of sessions exceed it.
(b) Cache-adjusted token count versus $L$, with one point per session and colors indicating task families.
The inset compares the median per-session ratios $U_{\mathrm{all}}/L$ and $U_{\mathrm{cache}}/L$ before and after subtracting cached input tokens ($23\times$ and $3.8\times$, respectively).}
\label{fig:trajectory-scale-overhead}
\end{figure}

We collect 1,556 rollout sessions from Stage 1 (steps 1--14) to characterize the trajectory scale of long-horizon web development.
Because context compaction rewrites the interaction history, per-turn token counts alone do not determine the exact length of the full deduplicated transcript.
We therefore summarize session scale using the proxy $L = \max_t P_t + \sum_t O_t$, where $P_t$ and $O_t$ denote the prompt and completion token counts at interaction turn $t$, respectively.
The median $L$ is 364k tokens, the 90th percentile is 903k, and the maximum reaches 2.02M.
As shown in Figure~\ref{fig:trajectory-scale-overhead}a, 97.9\% of sessions exceed the 64k training window under this measure.
This mismatch is also evident without the prompt-length term: cumulative generated tokens alone, $\sum_t O_t$, exceed the window in 94.5\% of sessions.
These statistics motivate context management and the construction of multiple training samples from each long rollout.

\begin{figure}[t]
\centering
\includegraphics[width=\linewidth]{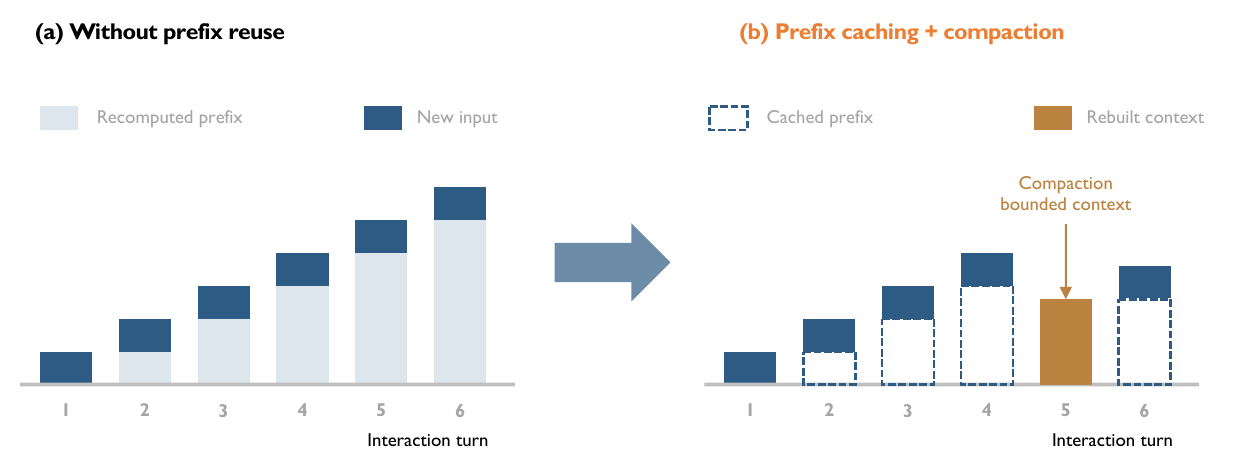}
\caption{\textbf{Context compaction and prefix caching.}
Left: historical inputs are repeatedly prefilled without prefix reuse.
Right: matching prefixes are cached, and compaction rebuilds a bounded context.
Filled regions indicate input requiring prefill; dashed outlines indicate reused prefixes.
The two panels illustrate mechanisms rather than a controlled timing comparison.
Turns and bar heights are schematic, not measured; decoding and cache-access costs are not depicted.}
\label{fig:compaction-caching}
\end{figure}

Long trajectories also incur repeated-input overhead during inference.
The median session spans 153 interaction turns, with successive requests repeatedly including retained history.
Let $C_t$ denote the cached prompt-token count.
We compare total requested tokens, $U_{\mathrm{all}} = \sum_t (P_t + O_t)$, with cache-adjusted tokens, $U_{\mathrm{cache}} = \sum_t (P_t - C_t + O_t)$.
The median per-session ratios $U_{\mathrm{all}}/L$ and $U_{\mathrm{cache}}/L$ are 23 and 3.8, respectively (Figure~\ref{fig:trajectory-scale-overhead}b).
Prefix caching reuses matching history prefixes, while compaction limits the context retained across interactions (Figure~\ref{fig:compaction-caching}).
These mechanisms address repeated-prefill overhead and context growth, respectively.

\section{Verifier Reliability}
\label{app:verifier}

Two model-based verifiers are used in this work. The \emph{reward judge} scores the final artifact of every rollout during reinforcement learning and forms the vision--language-model leg of the terminal reward. The \emph{acceptance judge} operates the running application at the end of each Webdev-Bench session and assigns every item of a hidden checklist an integer grade from $0$ to $3$, with $3$ denoting full satisfaction. This appendix measures the repeated-grading error of both and shows, on concrete cases, that their scores correspond to visible differences in the artifacts. Notation follows Appendix~\ref{sec:protocol}: a requirement counts as satisfied only when all of its items receive grade $3$, and the mean over requirements is the requirement satisfaction score (RS); restricting items to the function-and-coverage, interaction, and visual dimensions gives FS, IS, and VS.

\subsection{Training Reward Judge: Self-Consistency and Signal-to-Noise}
\label{app:verifier-reward}

The reward judge (GLM-5V-Turbo~\citep{hong2026glm}) grades a rendered screenshot on a five-level anchored ordinal scale per dimension (excellent, good, fair, poor, broken, mapped to $1.0$, $0.75$, $0.5$, $0.25$, $0$). The dimensions are fixed by task type---visual fidelity, content coverage, layout, text match, and render health for reproduction tasks; design quality, content quality, functionality, and render health for generation tasks---and their weighted sum is the judge-leg score $f_{\mathrm{vlm}}\in[0,1]$. The terminal reward is $R=g\cdot\big(w_{\mathrm{det}}f_{\mathrm{det}}+w_{\mathrm{vlm}}f_{\mathrm{vlm}}\big)$, where $g$ is the product of the build, entry, and render gates, $f_{\mathrm{det}}$ is the programmatic deterministic leg (pixel comparison for reproduction, DOM-structure checks for generation), and $w_{\mathrm{vlm}}=0.4$.

Grading the same rendered screenshot repeatedly gives a per-draw standard deviation of $0.074$ on the unit score range. The median of three draws reduces this to $0.019$, and five draws lower it by only a further $7\%$. Training uses the median of three throughout, which bounds the judge leg's measurement error at about $2\%$ of the range.

Whether this error is small enough depends on the differences it must resolve. GRPO builds its gradient from relative advantages among rollouts of the same task, so the relevant comparison is the within-group spread. In the Stage 1 training groups themselves (eight rollouts per task; $220$ groups and $1{,}280$ gate-passing rollouts from steps 1--14), the pooled within-group standard deviation of the reward is $0.078$, against a judge noise of roughly $0.008$ once propagated through its weight in the reward: genuine within-group differences are about ten times the measurement noise, and the gradient is driven by policy differences rather than grading jitter.

\begin{figure}[t]
\centering
\includegraphics[width=\linewidth]{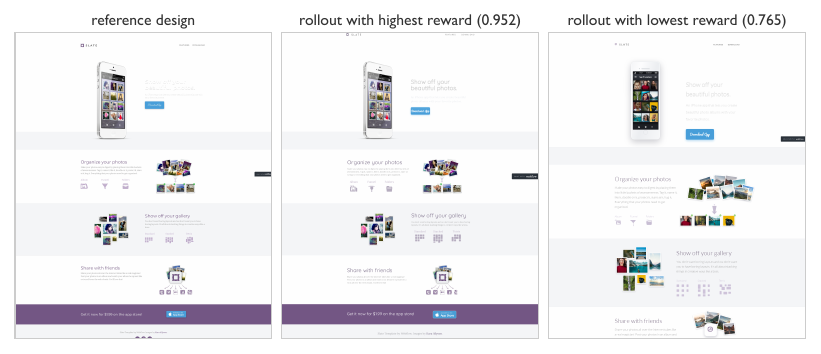}
\caption{\textbf{Training-side case: renders of the highest- and lowest-reward rollouts of one reproduction task.} Panels show the reference design, the rollout with reward $0.952$, and the rollout with reward $0.765$, each cropped to the same height from the top of the page and scaled at native aspect ratio.}
\label{fig:verifier-traindemo}
\end{figure}

What a within-group spread corresponds to in the artifacts can be seen from the highest- and lowest-reward rollouts of a reproduction task with a wide within-group range (Figure~\ref{fig:verifier-traindemo}, Table~\ref{tab:verifier-traindemo}). The highest-reward artifact matches the reference in layout, section order, and the horizontal placement of the phone mock-up; the judge gives $1.0$ on both visual fidelity and layout, and the pixel leg scores $0.920$. The lowest-reward artifact turns the mock-up into a tall vertical image, widens section spacing, and stretches the page by about $18\%$; the judge lowers visual fidelity to $0.75$ and the pixel leg falls to $0.659$. Both legs move in the same direction, and the judge's higher score corresponds to a visibly closer reproduction.

\begin{table}[t]
\centering
\small
\caption{\textbf{Judge dimension scores and deterministic-leg score for the two rollouts in Figure~\ref{fig:verifier-traindemo}.} Dimension scores are medians of three draws.}
\label{tab:verifier-traindemo}
\setlength{\tabcolsep}{4.5pt}
\begin{tabular}{lccccccc}
\toprule
Rollout & Visual fidelity & Coverage & Layout & Text match & Render health & Pixel leg $f_{\mathrm{det}}$ & Reward $R$ \\
\midrule
Highest & 1.00 & 1.00 & 1.00 & 1.00 & 1.00 & 0.920 & 0.952 \\
Lowest & 0.75 & 1.00 & 1.00 & 1.00 & 1.00 & 0.659 & 0.765 \\
\bottomrule
\end{tabular}
\end{table}

\Needspace*{20\baselineskip}
\subsection{Webdev-Bench Acceptance Judge: Repeated-Grading Consistency}
\label{app:verifier-acceptance}

\begin{wrapfigure}{r}{0.5\linewidth}
    \centering
    \vspace{-6pt}
    \includegraphics[width=\linewidth]{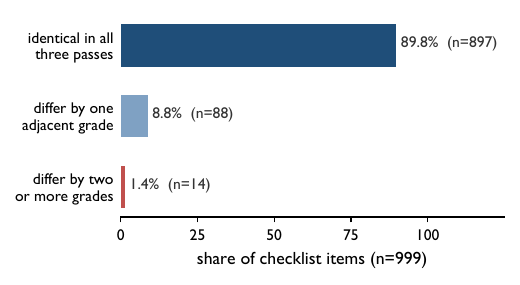}
    \caption{
        \textbf{Agreement of the acceptance judge across three independent passes.}
        Share of 999 checklist items whose three grades are identical, differ by one adjacent grade, or differ by two or more grades.
    }
    \label{fig:verifier-agree}
    \vspace{-8pt}
\end{wrapfigure}

For the 77 tasks from six model runs on which three independent terminal acceptance passes were executed and all three per-item outputs parse, we align the three grades of every item, giving 999 item triples; each pass is a separate session in which the judge restarts and operates the application before grading. At the item level, each item is classified by the range of its three grades: identical in all passes, differing by at most one grade, or differing by two or more. At the task level, RS is computed for each pass and compared across the three passes. Figure~\ref{fig:verifier-agree} shows the result: $89.8\%$ of items receive identical grades in all three passes, $8.8\%$ differ by one grade, and $1.4\%$ by two or more. The three passes agree closely at both the item-grade and the task-level RS layer, indicating that the acceptance judge's grading is stable and reproducible.

The three-pass record of one task shows what this agreement looks like in practice (Table~\ref{tab:verifier-acceptdemo}). The task is a store-operations dashboard with four requirements and 22 items, built by Claude Opus 4.8. The number of items graded $3$ under each requirement is identical across the three passes, and the same item is the one deducted every time: requirement three asks that, once six stores are on the watch list, clicking a further store shows a visible message that the limit has been reached. In all three passes the judge filled the list to six, confirmed that the seventh click was blocked, and recorded that no visible message appeared, assigning grade $2$ each time. Three independent sessions deducting at the same place for the same reason indicates that the verdict reflects the artifact rather than grading randomness.

\begin{table}[t]
\centering
\small
\caption{\textbf{Evaluation-side consistency case: per-requirement outcome of three acceptance passes on the same artifact.} Cells give the number of items graded $3$ over the number of items in the requirement.}
\label{tab:verifier-acceptdemo}
\begin{tabular}{lcccc}
\toprule
& Req.\ 1 dashboard & Req.\ 2 search and filters & Req.\ 3 watch list & Req.\ 4 sidebar and views \\
\midrule
Pass 1 & 5/5 & 6/6 & 5/6 & 5/5 \\
Pass 2 & 5/5 & 6/6 & 5/6 & 5/5 \\
Pass 3 & 5/5 & 6/6 & 5/6 & 5/5 \\
\bottomrule
\end{tabular}
\end{table}

\begin{figure}[t]
\centering
\includegraphics[width=\linewidth]{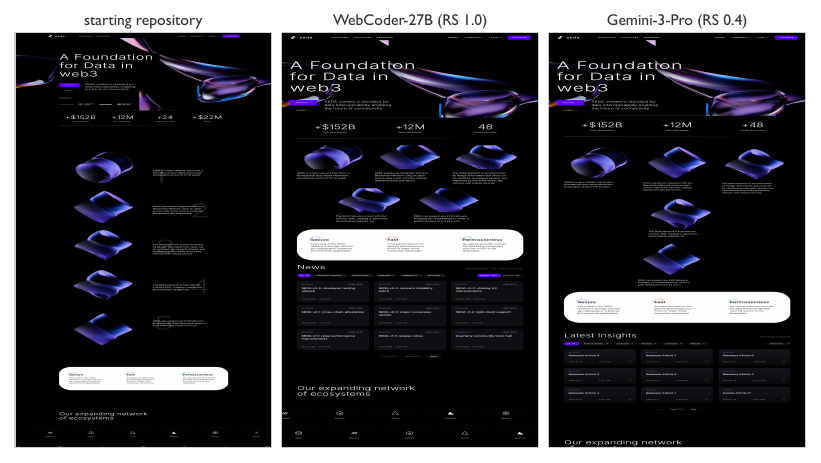}
\caption{\textbf{Evaluation-side contrast case: final-state screenshots of the same task built by two model lines.} Panels show the starting repository's homepage, the \ModelName{}-27B artifact, and the Gemini-3-Pro artifact, each cropped to the top of the page at the same aspect ratio. \ModelName{} compacts the five ``Why SEDA'' parts into two rows of three and two; the Gemini-3-Pro artifact leaves the last two parts staggered below the first row, exceeding the height budget, with the numerals hidden behind the images.}
\label{fig:verifier-contrast}
\end{figure}

Whether a deduction corresponds to a visible defect is checked on one task built by two model lines (Figure~\ref{fig:verifier-contrast}, Table~\ref{tab:verifier-contrast}). The task is c22 of Section~\ref{sec:case-studies}: starting from the website of a blockchain project, the agent repairs dead navigation links, tightens the hero, compacts the five ``Why SEDA'' parts into two rows of three and two within a 1400-pixel height budget, adds a 48-article newsroom, and wires it into the navigation and the statistics bar. The \ModelName{}-27B artifact passes all 27 checklist items (RS $1.0$). The Gemini-3-Pro artifact completes the first two rounds and the newsroom, but its five parts remain staggered---three in the first row and the remaining two offset one below the other---the block is 1613 pixels tall, and the numerals 1--5 are hidden behind the images. The judge records exactly these observations, measuring the card offsets (451 pixels between the fourth and fifth cards) and the block height and confirming by pixel comparison that the numerals are covered, and grades the layout and numeral items $1$ and the height item $2$; two further items are graded below $3$ because the statistics cell is not a link and the newsroom filter pills do not follow the site's button style. Requirements three, four, and five are therefore unsatisfied (RS $0.4$), and the two deductions that decide requirement three are visible in the screenshot.

\begin{table}[t]
\centering
\small
\caption{\textbf{Per-requirement acceptance outcome and items below grade $3$ for the two artifacts in Figure~\ref{fig:verifier-contrast}.}}
\label{tab:verifier-contrast}
\begin{tabular}{lccp{6.6cm}}
\toprule
Model & Requirements 1--5 & RS & Items below grade 3 (grade) \\
\midrule
\ModelName{}-27B & 1, 1, 1, 1, 1 & 1.0 & none \\
Gemini-3-Pro & 1, 1, 0, 0, 0 & 0.4 & five parts in two rows (1), numerals visible with their texts (1), block height within 1400\,px (2); pager shown for single-page categories (2); statistics cell linked to the newsroom (1), button style consistency (2) \\
\bottomrule
\end{tabular}
\end{table}





\section{Analysis of Agent Behavior}
\label{sec:agent-behavior}


To analyze the development behavior associated with changes in training reward, we collect 1,556 rollouts from Stage 1, steps 1--14, and examine their model-call records, saved conversation histories, and final-artifact evaluations.
Based on the reward curve, we divide the analysis into three windows: Early (steps 1--4), Trough (steps 8--11), and Recovery (steps 12--14).
The behavioral analysis focuses on replication tasks, where the reward decline is more pronounced.
The three windows contain 291, 267, and 181 replication rollouts spanning 30, 27, and 28 tasks, respectively.

We measure three aspects of tool use from the observed conversation windows after deduplicating overlapping history:
\begin{itemize}
    \item \textbf{Redundant operations.}
    The fraction of observed tool calls that repeat the same tool and arguments without an intervening call classified as modifying.
    \item \textbf{Observed skill use.}
    The fraction of rollouts with observable text that contain at least one skill invocation.
    \item \textbf{Observed webshot use.}
    The fraction of rollouts with observable text that contain at least one \texttt{webshot} command.
\end{itemize}

\begin{figure}[h]
\centering
\includegraphics[width=\linewidth]{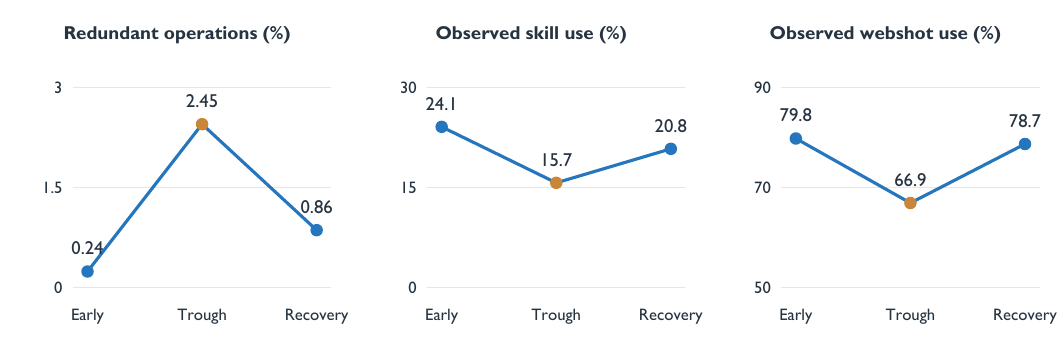}
\caption{
Tool-use patterns across training windows in replication tasks:redundant-operation rate, observed skill use, and observed webshot use.
Early, Trough, and Recovery correspond to steps 1--4, 8--11, and 12--14, respectively.
}
\label{fig:behavioral-changes}
\end{figure}

Figure~\ref{fig:behavioral-changes} shows behavioral changes across the three windows for replication tasks.
During the trough, the redundant-operation rate rises from 0.24\% to 2.45\%, while skill invocation falls from 24.1\% to 15.7\% and webshot invocation from 79.8\% to 66.9\%.
The reward decline thus coincides with a larger fraction of tool calls repeating previous operations and fewer rollouts invoking Skills or webshot.
During recovery, the redundant-operation rate decreases to 0.86\%, while skill and webshot invocation rise to 20.8\% and 78.7\%, respectively.
All three metrics move toward their early levels, suggesting a partial improvement in development behavior alongside reward recovery, with fewer redundant operations and renewed use of Skills and webshot.
We further examine these changes through policy entropy, generated-word frequencies, and an observed operation sequence.

\begin{figure}[h]
\centering
\includegraphics[width=\linewidth]{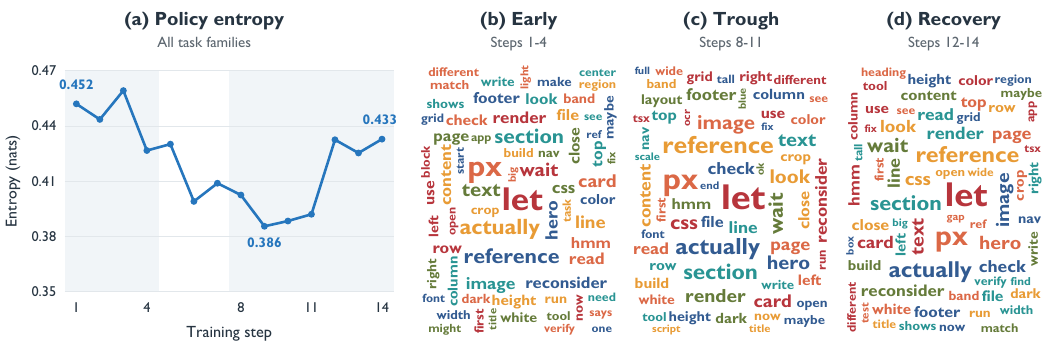}
\caption{
(a) Training-policy entropy.
(b--d) Frequent words in replication tasks during Early, Trough, and Recovery.
Font sizes use a shared square-root mapping of word frequency across all three clouds; color does not encode a statistic.
}
\label{fig:entropy-word-cloud}
\end{figure}

To examine whether these behavioral changes coincide with changes in the model's predictive distribution, we analyze policy entropy during training.
Entropy is computed over the full vocabulary and averaged over generated tokens in each training batch.
As shown in Figure~\ref{fig:entropy-word-cloud}(a), policy entropy decreases from 0.452 nats at step 1 to 0.386 at step 9, then rises to 0.433 at step 14.
Lower entropy during the trough reflects more concentrated predictions on the observed contexts and coincides with increased redundant operations.
During recovery, entropy rises as redundant operations decrease.

We further analyze generated language in replication trajectories to examine how the model's descriptions of development change during training.
The corpus is drawn from deduplicated observed conversations, retaining reasoning text and prose outside tool calls.
We filter out code, tool arguments, stopwords, and task-specific terms, and merge word forms.
Frequencies are normalized per 10,000 retained words within each window.
Figure~\ref{fig:entropy-word-cloud}(b--d) shows the frequent-word distributions, and Table~\ref{tab:word-frequencies} reports frequencies for selected terms.

\begin{table}[h]
    \centering
    \caption{Selected word frequencies per 10,000 retained words.}
    \label{tab:word-frequencies}
    \small
    \renewcommand{\arraystretch}{1.15}
    \begin{tabular*}{\linewidth}
        {@{\extracolsep{\fill}}lccc@{}}
        \toprule
        \textbf{Word} & \textbf{Early} &
        \textbf{Trough} & \textbf{Recovery} \\
        \midrule
        \texttt{look}      & 63.2  & 125.3 & 89.1  \\
        \texttt{reference} & 148.5 & 219.3 & 183.3 \\
        \texttt{check}     & 60.1  & 85.3  & 84.3  \\
        \texttt{edit}      & 19.3  & 8.1   & 10.3  \\
        \texttt{run}       & 42.2  & 31.1  & 37.6  \\
        \bottomrule
    \end{tabular*}
\end{table}

During the trough, \texttt{check} increases from 60.1 to 85.3 occurrences per 10,000 words, and \texttt{reference} increases from 148.5 to 219.3, while webshot invocation declines.
Inspection-related language becomes more frequent without a corresponding increase in webshot use.
This divergence shows that emphasis on inspection in generated text and actual tool use can change in different directions.
Operation records are therefore needed to determine whether these descriptions translate into development actions.

\begin{figure}[h]
\centering
\includegraphics[width=\linewidth]{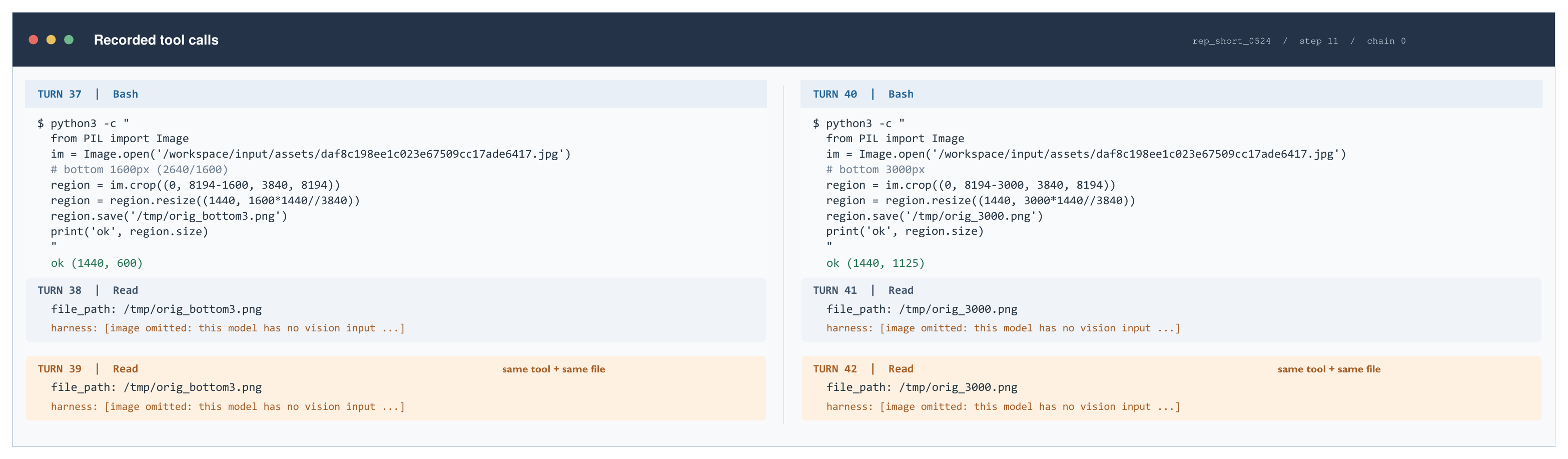}
\caption{
\textbf{A terminal-style excerpt from a recorded rollout.}
Commands, paths, outputs, and image-omission notices are reproduced from the saved records; orange highlights repeated reads.
\texttt{Read} is an agent tool, and the \texttt{harness} lines contain framework feedback.
The figure is a typeset transcript, not an original terminal screenshot.
}
\label{fig:terminal-case}
\end{figure}

To illustrate repetitive behavior during the trough, we select a highly repetitive replication rollout from training step 11.
Figure~\ref{fig:terminal-case} presents turns 37--42 from a single continuous message chain: after cropping the reference image, the model reads the same file twice, then crops a larger region and again reads the resulting file twice.

The cropping commands successfully create files, but every \texttt{Read} call returns an image-omission notice, so the image content is not passed to the policy.
No repository edits are observed in the saved windows, and the final artifact retains scaffold content, receiving a task reward of 0.031.
This case illustrates that repeated calls to inspection tools do not necessarily provide usable feedback or advance the project.
Assessing the contribution of long-horizon interactions therefore requires examining whether the operations advance the artifact, motivating segment-level credit assignment based on artifact progress.

\section{Full Results}
\label{app:full-results}

Table~\ref{tab:full-wb} reports the complete Webdev-Bench metrics of every evaluated model, using the same models, ordering, and scoring as Table~\ref{tab:main_results}. The four right-hand columns give the item-level grade-3 rate per requirement dimension, which is unaffected by the conjunctive rule and shows the absolute degree of completion: \ModelName{} reaches 96\% on interaction items, the highest of all models, and 85\% on functional items, against 89\% and 79\% for the base model.

Table~\ref{tab:full-wb-type} splits the results by task starting point. The gains of \ModelName{} over the base model concentrate on text-to-project tasks (RS 70.0 versus 57.2, with Claude-Opus-4.8 at 60.0) and on replication tasks (RS 58.9 versus 51.7); on text-to-project tasks \ModelName{} has the highest value in all four metrics.

\begin{table}[t]
    \centering
    \small
    \renewcommand{\arraystretch}{1.15}
    \setlength{\tabcolsep}{5pt}
    \caption{\textbf{Full Webdev-Bench results.} The left six columns follow the metrics and ordering of Table~\ref{tab:main_results}; the right four columns give the item-level grade-3 rate, i.e., the fraction of checklist items of each dimension graded $3$ at terminal acceptance (items of undelivered tasks count as grade $0$). Best in bold, second best underlined.}
    \label{tab:full-wb}
    \resizebox{\linewidth}{!}{
    \begin{tabular}{lcccccc|cccc}
    \toprule
    \multirow{2}{*}{\textbf{Model}} & \multicolumn{6}{c|}{\textbf{Webdev-Bench}} & \multicolumn{4}{c}{\textbf{Item grade-3 rate}} \\
    \cmidrule(lr){2-7}\cmidrule(lr){8-11}
    & \textbf{RS} & \textbf{VS} & \textbf{FS} & \textbf{IS} & \textit{Avg} & \textit{DSR} & Functional & Coverage & Interaction & Visual \\
    \midrule
    \rowcolor{gray!12}\multicolumn{11}{c}{\textit{\textbf{Proprietary Models}}} \\
    Claude-Opus-4.8 & \ranksecond{65.1} & 75.0 & \ranksecond{70.7} & 84.1 & \ranksecond{73.7} & 100.0\% & \ranksecond{83.6} & 83.3 & \ranksecond{91.6} & 86.1 \\
    GPT-5.4 & 59.6 & 72.7 & 70.0 & 81.0 & 70.8 & 100.0\% & 83.1 & \ranksecond{85.2} & 91.3 & 84.8 \\
    GPT-5 & 53.5 & 64.0 & 65.8 & 74.6 & 64.5 & 100.0\% & 81.4 & 80.6 & 86.3 & 77.0 \\
    Gemini-3-Pro & 48.0 & 65.8 & 58.5 & 68.5 & 60.2 & 100.0\% & 75.2 & 75.9 & 80.3 & 77.7 \\
    Gemini-3-Flash & 56.6 & 65.8 & 64.8 & \ranksecond{85.0} & 68.1 & 95.0\% & 80.4 & 84.3 & \ranksecond{91.6} & 78.6 \\
    Seed-1.8 & 52.6 & 66.5 & 63.8 & 71.5 & 63.6 & 100.0\% & 79.2 & 84.3 & 87.0 & 79.9 \\
    \rowcolor{gray!12}\multicolumn{11}{c}{\textit{\textbf{Open-Source Models}}} \\
    Kimi-K2.5 & 46.8 & 61.5 & 55.0 & 67.7 & 57.7 & 82.5\% & 68.4 & 72.2 & 73.6 & 68.3 \\
    MiMo-V2.5 & 14.9 & 25.2 & 21.2 & 22.6 & 21.0 & 65.0\% & 28.4 & 46.3 & 27.8 & 37.5 \\
    Step-3.7-Flash & 10.8 & 20.0 & 14.0 & 19.0 & 15.9 & 60.0\% & 18.6 & 34.3 & 23.1 & 28.2 \\
    Qwen3.8-27B & 64.8 & \ranksecond{77.9} & 68.8 & 79.8 & 72.8 & 95.0\% & 79.2 & \ranksecond{85.2} & 88.6 & \ranksecond{86.7} \\
    Qwen3.5-397B-A17B & 28.7 & 39.0 & 31.9 & 36.5 & 34.0 & 80.0\% & 42.2 & 56.5 & 47.2 & 49.2 \\
    Qwen3.5-122B-A10B & 24.6 & 30.4 & 28.9 & 33.1 & 29.3 & 72.5\% & 36.3 & 44.4 & 40.8 & 41.7 \\
    Qwen3-VL-32B-Instruct & 2.5 & 6.3 & 5.4 & 4.4 & 4.6 & 50.0\% & 11.8 & 6.5 & 10.4 & 15.5 \\
    Qwen3-VL-8B-Instruct & 0.0 & 0.8 & 0.0 & 0.0 & 0.2 & 30.0\% & 0.5 & 1.9 & 1.0 & 1.9 \\
    \midrule
    \textbf{\ModelName{}-27B} & \rankbest{72.8} & \rankbest{81.9} & \rankbest{73.8} & \rankbest{90.8} & \rankbest{79.8} & 100.0\% & \rankbest{84.8} & \rankbest{88.0} & \rankbest{95.7} & \rankbest{89.3} \\
    \bottomrule
    \end{tabular}}
\end{table}

\begin{table}[t]
    \centering
    \small
    \renewcommand{\arraystretch}{1.15}
    \setlength{\tabcolsep}{4pt}
    \caption{\textbf{Webdev-Bench results by task starting point.} Tasks start from an existing repository (brownfield), a reference screenshot (replication), or a textual specification (text-to-project); each cell is the mean over the tasks of that type, with undelivered tasks counted as $0$. Best in bold, second best underlined.}
    \label{tab:full-wb-type}
    \resizebox{\linewidth}{!}{
    \begin{tabular}{lcccccccccccc}
    \toprule
    \multirow{2}{*}{\textbf{Model}} & \multicolumn{4}{c}{\textbf{Brownfield ($n$=11)}} & \multicolumn{4}{c}{\textbf{Replication ($n$=9)}} & \multicolumn{4}{c}{\textbf{Text-to-project ($n$=20)}} \\
    \cmidrule(lr){2-5}\cmidrule(lr){6-9}\cmidrule(lr){10-13}
    & \textbf{RS} & \textbf{VS} & \textbf{FS} & \textbf{IS} & \textbf{RS} & \textbf{VS} & \textbf{FS} & \textbf{IS} & \textbf{RS} & \textbf{VS} & \textbf{FS} & \textbf{IS} \\
    \midrule
    \rowcolor{gray!12}\multicolumn{13}{c}{\textit{\textbf{Proprietary Models}}} \\
    Claude-Opus-4.8 & \ranksecond{79.5} & \ranksecond{90.9} & 90.0 & 90.9 & \rankbest{58.9} & 61.1 & \rankbest{78.3} & \ranksecond{77.8} & 60.0 & 72.5 & 56.7 & 83.2 \\
    GPT-5.4 & 71.8 & 75.0 & 87.7 & \ranksecond{95.5} & \ranksecond{55.0} & \rankbest{71.3} & 66.7 & 74.1 & 55.0 & 72.1 & 61.7 & 76.1 \\
    GPT-5 & 59.5 & 65.9 & 86.8 & 74.2 & 37.2 & 50.9 & 46.7 & 61.1 & 57.5 & 68.8 & \ranksecond{62.9} & 80.8 \\
    Gemini-3-Pro & 55.0 & 62.1 & 80.5 & 85.6 & 23.3 & 52.8 & 32.2 & 40.7 & 55.3 & \ranksecond{73.8} & 58.3 & 71.6 \\
    Gemini-3-Flash & 58.2 & 65.2 & 85.5 & 90.2 & 44.4 & 66.7 & 55.6 & \rankbest{79.6} & \ranksecond{61.3} & 65.8 & 57.5 & \ranksecond{84.6} \\
    Seed-1.8 & 52.3 & 62.1 & 80.5 & 66.7 & 42.8 & 63.9 & 55.6 & 55.6 & 57.3 & 70.0 & 58.3 & 81.3 \\
    \rowcolor{gray!12}\multicolumn{13}{c}{\textit{\textbf{Open-Source Models}}} \\
    Kimi-K2.5 & 60.0 & 76.5 & 82.3 & 83.3 & 32.8 & 53.7 & 50.6 & 55.6 & 45.8 & 56.7 & 42.1 & 64.6 \\
    MiMo-V2.5 & 23.2 & 43.2 & 29.1 & 34.8 & 2.2 & 8.3 & 24.4 & 13.0 & 16.0 & 22.9 & 15.4 & 20.2 \\
    Step-3.7-Flash & 15.5 & 34.1 & 21.8 & 27.3 & 0.0 & 5.6 & 7.8 & 3.7 & 13.0 & 18.8 & 12.5 & 21.3 \\
    Qwen3.8-27B & \rankbest{89.1} & \rankbest{97.7} & \rankbest{97.7} & 90.2 & 51.7 & \ranksecond{70.4} & 66.7 & 64.8 & 57.3 & 70.4 & 53.8 & 80.8 \\
    Qwen3.5-397B-A17B & 32.7 & 47.7 & 41.8 & 37.9 & 17.2 & 22.2 & 30.6 & 18.5 & 31.8 & 41.7 & 27.1 & 43.8 \\
    Qwen3.5-122B-A10B & 32.7 & 45.5 & 42.7 & 47.7 & 8.9 & 9.3 & 15.0 & 3.7 & 27.3 & 31.7 & 27.5 & 38.3 \\
    Qwen3-VL-32B-Instruct & 0.0 & 0.0 & 0.0 & 0.0 & 0.0 & 0.0 & 0.0 & 0.0 & 5.0 & 12.5 & 10.8 & 8.8 \\
    Qwen3-VL-8B-Instruct & 0.0 & 0.0 & 0.0 & 0.0 & 0.0 & 0.0 & 0.0 & 0.0 & 0.0 & 1.7 & 0.0 & 0.0 \\
    \midrule
    \textbf{\ModelName{}-27B} & \rankbest{89.1} & \rankbest{97.7} & \ranksecond{93.2} & \rankbest{97.0} & \rankbest{58.9} & 68.5 & \ranksecond{70.6} & 74.1 & \rankbest{70.0} & \rankbest{79.2} & \rankbest{64.6} & \rankbest{95.0} \\
    \bottomrule
    \end{tabular}}
\end{table}

\section{Human Evaluation}
\label{app:human-eval}

The acceptance judge is itself a model, so we ask professional annotators with front-end development experience to check, by pairwise comparison on the 40-task Webdev-Bench suite, whether its preference order agrees with theirs. Each item pairs the final deliverable of \ModelName{}-27B with that of one of four opponents on the same task: the Qwen3.8-27B base model, Claude Opus 4.8, GPT-5.4, and Gemini-3-Flash, giving $40\times 4=160$ items. The two deliverables are shown anonymously side by side as columns A and B, with the column-to-model assignment drawn at random for every item; each column shows the full-page screenshot captured at the terminal acceptance pass and the per-round screenshots, and the task header lists the requirements exactly as they were revealed to the model together with the starting material. The annotator chooses Win~A, Tie, or Win~B.

\Needspace*{16\baselineskip}
\begin{wrapfigure}{r}{0.5\linewidth}
    \centering
    \vspace{-8pt}
    \includegraphics[width=\linewidth]{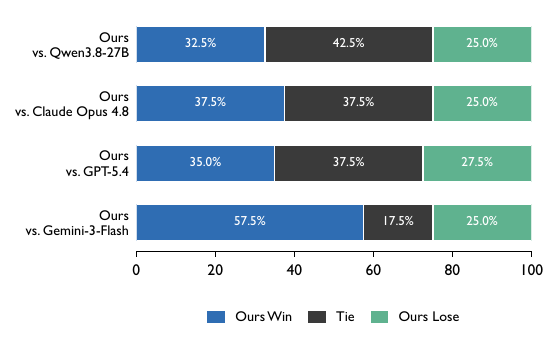}
    \caption{
        \textbf{Pairwise human preference for \ModelName{}-27B against four opponents on the 40-task Webdev-Bench suite.}
        Each row gives the share of the 40 items in which \ModelName{} wins, ties, or loses.
    }
    \label{fig:human-eval}
    \vspace{-8pt}
\end{wrapfigure}

The evaluation policy is requirement-driven: for every requirement the annotator checks whether the required content is present, whether functional results are consistent with the specification, and whether interaction state is visibly evidenced; visual fidelity and overall usability are considered only afterwards. A deliverable that fails to start loses outright, and a tie is reserved for pairs that satisfy essentially the same requirements with no clear visual gap.

Figure~\ref{fig:human-eval} reports the results. Against all four opponents the win-or-tie share of \ModelName{}-27B lies between $72.5\%$ and $75\%$: against the Qwen3.8-27B base it wins $32.5\%$, ties $42.5\%$, and loses $25.0\%$ of the items; against Claude Opus 4.8 it wins $37.5\%$, ties $37.5\%$, and loses $25.0\%$; against GPT-5.4 it wins $35.0\%$, ties $37.5\%$, and loses $27.5\%$; against Gemini-3-Flash it wins $57.5\%$, ties $17.5\%$, and loses $25.0\%$. Annotator comments show that disagreements concentrate on interaction requirements, such as whether a filter narrows the list correctly or whether an over-limit action is blocked with a message; these items can only be confirmed by operating the application, which is exactly what the acceptance judge verifies item by item on the running deliverable.


\section{Additional Case Studies}
\label{app:case-studies}

This appendix complements Section~\ref{sec:case-studies} with three views of the same two benchmarks: how \ModelName{} compares with other model lines on a task that all of them completed, what its full trajectory looks like on the two tasks of the main text, and a task on which its deliverable does not pass acceptance.

\subsection{Comparison with Baselines on a Shared Task}
\label{app:case-compare}

\begin{figure}[t]
\centering
\includegraphics[width=\linewidth]{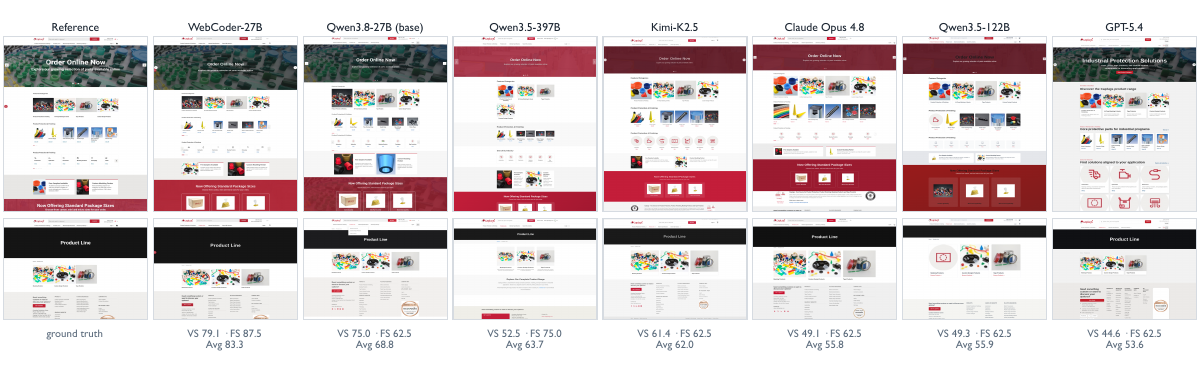}
\caption{\textbf{Vision2Web, caplugs: reference and final renders of seven model lines.} Top row shows the homepage, bottom row the product-catalogue page, both cropped from the top. Scores below each column are VS, FS, and their average for the whole six-page title.}
\label{fig:case-v2w-gallery}
\end{figure}

Figure~\ref{fig:case-v2w-gallery} shows the Vision2Web title \emph{caplugs}, on which every evaluated line delivered a running site, so the comparison is between complete deliverables rather than between delivery and failure. \ModelName{} leads on both metrics (VS $79.1$, FS $87.5$, Avg $83.3$). The base model keeps the layout but fails three of eight functional workflows; several larger lines replace the photographic hero with a flat colour band or alter the header and product grid of the catalogue page. The difference is not one of delivery rate: on this title \ModelName{} is ahead by $14.5$ points over the second-best line with every deliverable running.

\begin{figure}[t]
\centering
\includegraphics[width=\linewidth]{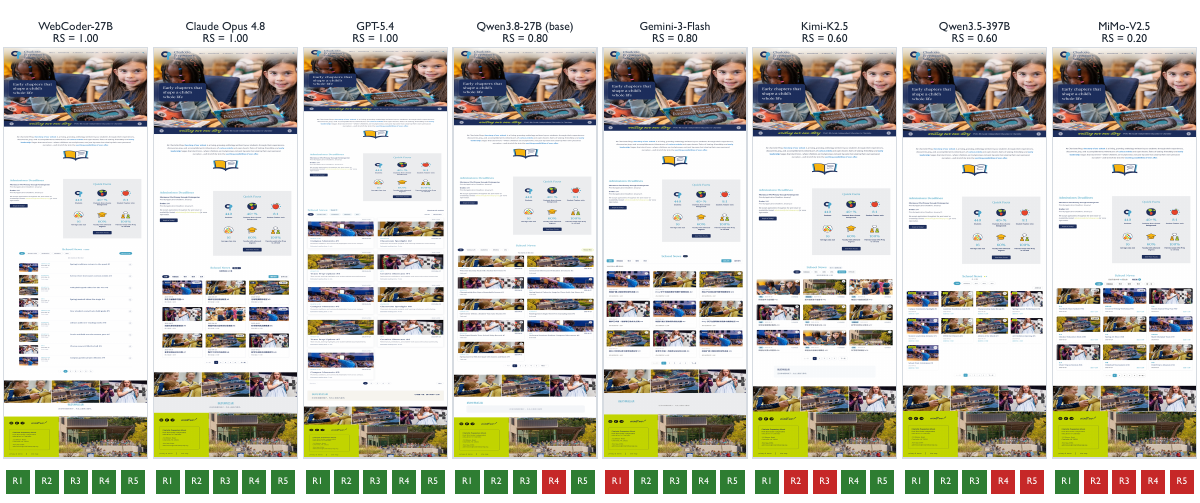}
\caption{\textbf{Webdev-Bench, c03: final applications of eight model lines and their per-requirement acceptance.} Each column shows the full final page after the fifth requirement; the strip below marks each requirement as satisfied (green, all items grade $3$) or not (red).}
\label{fig:case-wb-gallery}
\end{figure}

Figure~\ref{fig:case-wb-gallery} shows the Webdev-Bench task c03, a five-requirement modification of an existing school website, on which thirteen of the fifteen lines delivered a final application. The final pages look nearly identical because the starting repository is shared, yet conjunctive acceptance spreads the lines from RS $1.0$ to $0.2$. \ModelName{}, Claude Opus 4.8, and GPT-5.4 satisfy all five requirements. The base model and Gemini-3-Flash each miss one requirement: the base model fails the interaction item that asks for an empty-state message and a zero counter after all saved articles are removed, and Gemini-3-Flash fails a layout item of requirement one, leaving a blank gap where the promotional card was removed. The weaker lines lose two to four requirements although their pages contain the same sections. The case illustrates why the benchmark grades by requirement rather than by appearance.

\subsection{Complete Trajectories of \ModelName{}}
\label{app:case-ours}

\begin{figure}[t]
\centering
\includegraphics[width=\linewidth]{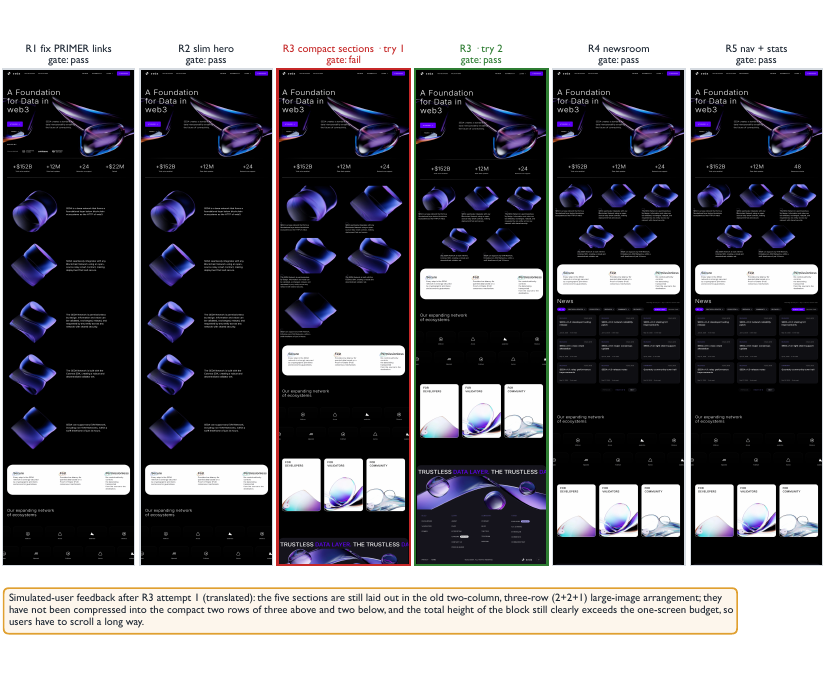}
\caption{\textbf{Webdev-Bench, c22: every attempt of \ModelName{}.} The first attempt at requirement three is rejected by the simulated-user gate (red frame); the second attempt passes (green frame) and the remaining rounds pass at the first attempt. The box gives the gate's feedback after the rejected attempt, translated from the original.}
\label{fig:case-wb-c22}
\end{figure}

Figure~\ref{fig:case-wb-c22} gives the full c22 trajectory summarised in the main text. The starting repository is the single-page website of SEDA. Requirement one repairs three dead PRIMER entries so that they scroll smoothly to their target block like the other links. Requirement two removes the investor row and the fundraising figure, tightens the hero from $1085$ to at most $900$ pixels, and redistributes the remaining statistics. Requirement three compacts the five staggered ``Why SEDA'' parts into two rows of three and two within $1400$ pixels while keeping every text and image. Requirement four adds a newsroom of $48$ articles in five categories with per-category counts on the filter buttons, two sort orders, nine articles per page with pagination and a page indicator, and a read-later mark with a counter and an empty-state message. Requirement five adds a Newsroom entry to the navigation that scrolls to the section and resets category and page while preserving the marks, and replaces one statistics cell with the article count. The simulated-user gate, which judges the current requirement's checklist items from the running page without revealing them, returns the feedback reproduced under the figure after the first attempt at requirement three: the five parts are still in the old layout and exceed the height budget. The second attempt rearranges them into two rows and passes. The other four rounds pass at the first attempt, and the terminal acceptance grades all $27$ checklist items $3$. The trajectory shows the intended division of labour between the two verifiers: the gate drives development with natural-language feedback and the terminal acceptance checks every item on the finished application.

\begin{figure}[t]
\centering
\includegraphics[width=\linewidth]{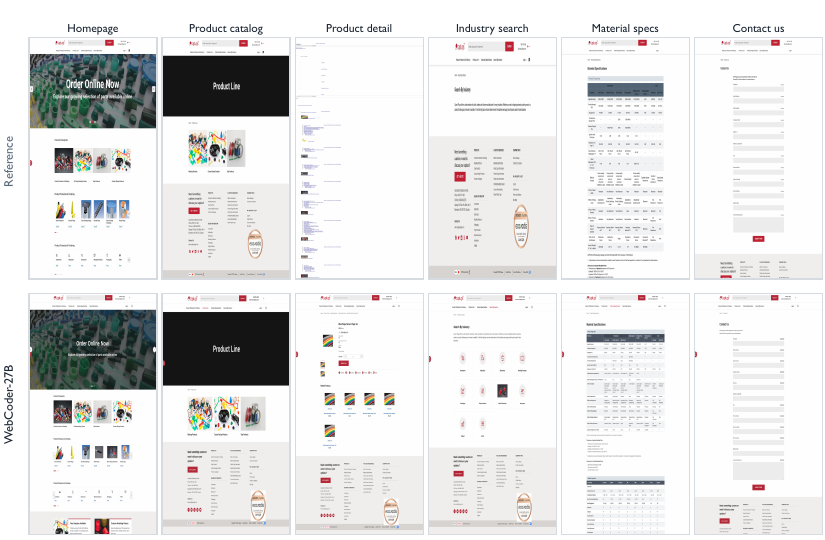}
\caption{\textbf{Vision2Web, caplugs: all six pages of the \ModelName{} deliverable.} Top row shows the reference screenshots, bottom row the judge-time renders of the served application, cropped from the top.}
\label{fig:case-v2w-ours}
\end{figure}

Figure~\ref{fig:case-v2w-ours} shows all six pages of the caplugs deliverable against their references. The homepage and catalogue pages score highest ($0.92$ and $0.95$ mean section score), and the material-specification and contact pages reproduce the dense table and the form; the industry-search page scores lowest because its search block and chat widget are missing, and the product-detail page is penalised for a header and share buttons that the cropped reference does not show. Of the thirteen functional test cases in six workflows, one fails: the footer link for an industry opens a page headed by the industry name instead of the expected product-line heading.

\subsection{A Trajectory That Does Not Pass Acceptance}
\label{app:case-fail}

\begin{figure}[t]
\centering
\includegraphics[width=\linewidth]{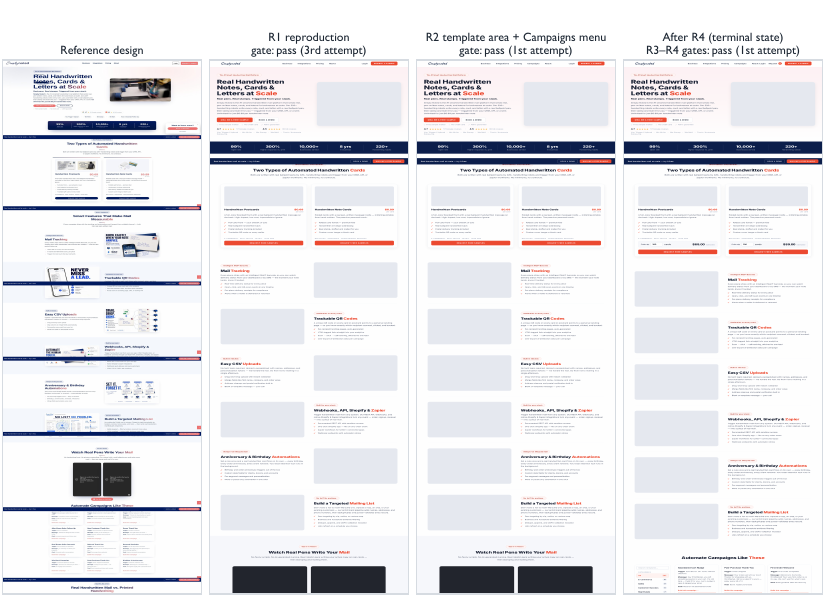}
\caption{\textbf{Webdev-Bench, c15: the \ModelName{} trajectory on a task graded RS $0.75$.} Panels show the reference design, the first passing state of requirement one (after two gate rejections), the state after requirement two, and the terminal state after requirement four, all cropped to the top $4{:}1$ region and shown at the same scale.}
\label{fig:case-wb-c15}
\end{figure}

Figure~\ref{fig:case-wb-c15} shows c15, which first reproduces the homepage of the handwritten-mail platform Simply Noted from a reference, then rearranges the template area into two columns with category counts, removes the video section and part of the statistics, and finally makes filtering, search, pagination, a plan counter, and a live price quote on two product cards work. The gate rejects the first two attempts at requirement one and passes the third; requirements two to four pass at the first attempt. Terminal acceptance grades $21$ of the $22$ checklist items $3$. The single deduction is a grade of $2$ on the price-quote item: for a postcard quantity of $499$ the page should show \$344.31 but renders ``\$344.30.999999999992724'', a floating-point remainder that the truncation logic does not cut off; the other four quantities specified by the item ($100$, $500$, $505$, and $800$ folded cards) are correct. This one formatting defect makes requirement four unsatisfied and leaves RS at $0.75$, which is the intended strictness of grading by requirement.

The same item is lost across model lines: on c15, twelve of the fourteen delivered lines fall below grade $3$ on the price quote, with Kimi-K2.5 and GPT-5 at grade $1$ and Qwen3.5-397B at $0$, while GPT-5.4, Gemini-3-Pro, Gemini-3-Flash, Seed-1.8, and \ModelName{} share grade $2$. Currency arithmetic with discount and truncation rules is where current agents most often fall one grade short on tasks of this kind; on c15 \ModelName{}'s RS of $0.75$ ties with Kimi-K2.5 and Seed-1.8 for the highest, with Claude Opus 4.8 and GPT-5.4 at $0.5$.

\clearpage

\section{Prompts and Skills}
\label{app:prompts}
This section presents the prompts and on-demand skills supplied for \ModelName{} training and Webdev-Bench task construction and evaluation.
We first organize the prompts by their role in each pipeline, then describe the skill library and reproduce its definitions.

\subsection{Prompts}
\paragraph{Rollout System Prompt.}
Recipient: development agent. Defines workspace constraints, skill selection, and the available rendering workflow.
  \begin{promptbox}{Prompt 1: Rollout System Prompt}
    \input{prompts/\stripexttrain_rollout_system_prompt.txt\relax}
  \end{promptbox}

\paragraph{First-Turn Task Envelope.}
Recipient: development agent. Combines task fields and the guidance blocks shown below; braces indicate runtime template fields.
  \begin{promptbox}{Prompt 2: First-Turn Task Envelope}
    \input{prompts/\stripexttrain_user_turn_envelope.txt\relax}
  \end{promptbox}

\paragraph{Generation Guidance.}
Recipient: development agent on text-only tasks. Directs an initial runnable implementation followed by screenshot-guided refinement.
  \begin{promptbox}{Prompt 3: Generation Guidance}
    \input{prompts/\stripexttrain_guidance_gen.txt\relax}
  \end{promptbox}

\paragraph{Replication Guidance.}
Recipient: development agent on screenshot tasks. Directs an initial full-page implementation followed by section-level visual comparison.
  \begin{promptbox}{Prompt 4: Replication Guidance}
    \input{prompts/\stripexttrain_guidance_rep.txt\relax}
  \end{promptbox}

\paragraph{Vision-Anchor Extraction.}
Recipient: vision model. Input: prepared reference views. Output: structured page identity, visual attributes, section order, and disambiguation cues.
  \begin{promptbox}{Prompt 5: Vision-Anchor Extraction}
    \input{prompts/\stripexttrain_vision_anchor_prompt.txt\relax}
  \end{promptbox}

\paragraph{Reference-Image Guidance.}
Recipient: development agent. Explains how to use the reference, the anchor file, and cropped views during reconstruction.
  \begin{promptbox}{Prompt 6: Reference-Image Guidance}
    \input{prompts/\stripexttrain_vision_section.txt\relax}
  \end{promptbox}

\paragraph{Build-and-Render Self-Check.}
Recipient: development agent. Uses webshot outputs to inspect rendering, build errors, missing text, and broken images.
  \begin{promptbox}{Prompt 7: Build-and-Render Self-Check}
    \input{prompts/\stripexttrain_webshot_section.txt\relax}
  \end{promptbox}

\paragraph{Long-Page Replication Example.}
A concrete task message with reference dimensions, content requirements, and delivery instructions. Task-asset paths identify inputs that are not included in this appendix.
  \begin{promptbox}{Prompt 8: Long-Page Replication Example}
    \input{prompts/\stripexttrain_rep_long.txt\relax}
  \end{promptbox}

\paragraph{Short-Page Replication Example.}
A concrete task message from the short-page replication family, retaining the supplied reference and project requirements.
  \begin{promptbox}{Prompt 9: Short-Page Replication Example}
    \input{prompts/\stripexttrain_rep_short.txt\relax}
  \end{promptbox}

\paragraph{Text-to-Project Example.}
A concrete generation task with textual requirements and no reference image.
  \begin{promptbox}{Prompt 10: Text-to-Project Example}
    \input{prompts/\stripexttrain_prd_gen.txt\relax}
  \end{promptbox}

\paragraph{Fidelity Judge.}
Input: reference image and rendered image. Output: anchored grades and reasons for reconstruction quality across the specified dimensions.
  \begin{promptbox}{Prompt 11: Fidelity Judge}
    \input{prompts/\stripexttrain_judge_fidelity.txt\relax}
  \end{promptbox}

\paragraph{Absolute Judge.}
Input: rendered image and task text when available. Output: anchored grades and reasons for design, content, visually inferred functionality, and rendering health.
  \begin{promptbox}{Prompt 12: Absolute Judge}
    \input{prompts/\stripexttrain_judge_absolute.txt\relax}
  \end{promptbox}

\paragraph{Reference-Image Screening.}
Input: candidate reference screenshot. Output: a suitability decision for use as a reconstruction target.
  \begin{promptbox}{Prompt 13: Reference-Image Screening}
    \input{prompts/\stripextwebdev_reference_image_qc.txt\relax}
  \end{promptbox}

\paragraph{Task Authoring.}
Input: task seed or blueprint and the available reference material. Output: development requirements and an associated acceptance checklist. Supplied revision annotations are retained.
  \begin{promptbox}{Prompt 14: Task Authoring}
    \input{prompts/\stripextwebdev_author_system.txt\relax}
  \end{promptbox}

\paragraph{Independent Task Review.}
Input: a proposed task and checklist. Output: a review decision and actionable issues, including checks on observability and requirement traceability.
  \begin{promptbox}{Prompt 15: Independent Task Review}
    \input{prompts/\stripextwebdev_reviewer_system.txt\relax}
  \end{promptbox}

\paragraph{Difficulty Revision.}
Input: an existing task. Output: a revised task with stronger state coupling or boundary cases, subject to the stated consistency constraints.
  \begin{promptbox}{Prompt 16: Difficulty Revision}
    \input{prompts/\stripextwebdev_hardening_system.txt\relax}
  \end{promptbox}

\paragraph{Controlled Defect Injection Example.}
A task-specific example for introducing a constrained defect. It is an example instance, not a universal injection template.
  \begin{promptbox}{Prompt 17: Controlled Defect Injection Example}
    \input{prompts/\stripextwebdev_saboteur_example.txt\relax}
  \end{promptbox}

\paragraph{Builder Prompt.}
Recipient: model under evaluation. Input: requirements and starting-project context. Output: a runnable implementation following the startup and delivery contract.
  \begin{promptbox}{Prompt 18: Builder Prompt}
    \input{prompts/\stripextwebdev_builder_prompt.txt\relax}
  \end{promptbox}

\paragraph{Acceptance Prompt.}
Recipient: acceptance agent. Input: the running application and acceptance checklist. Output: item-wise grades and evidence from browser observations and operations.
  \begin{promptbox}{Prompt 19: Acceptance Prompt}
    \input{prompts/\stripextwebdev_acceptance_prompt.txt\relax}
  \end{promptbox}

\subsection{On-Demand Skills}

The training agent has access to task-specific skills.
Screenshot replication is supported by seven skills covering
input analysis, planning, asset processing, implementation,
and validation, while text-only generation uses a dedicated
text-to-project skill.
Table~\ref{tab:skill-overview} summarizes their roles.

\begin{table}[t]
\centering
\caption{Overview of the skills available to the training agent.
Names are shortened for readability.}
\label{tab:skill-overview}
\small
\renewcommand{\arraystretch}{1.15}
\setlength{\tabcolsep}{4pt}
\begin{tabularx}{\linewidth}{
  @{}l >{\raggedright\arraybackslash}X@{}
}
\toprule
\textbf{Skill} & \textbf{Role} \\
\midrule
\multicolumn{2}{@{}l}{\textit{Screenshot replication}} \\
\addlinespace[3pt]
Pipeline coordination &
Coordinates planning, implementation, and validation. \\

Input analysis &
Interprets reference inputs and identifies the task mode. \\

Long-image planning &
Divides tall screenshots into semantic sections. \\

Flow planning &
Models multiple reference views as application states
and transitions. \\

Asset processing &
Prepares visual assets and selects their representations. \\

Code generation &
Implements and integrates the planned frontend. \\

Validation and repair &
Checks build and rendering results and repairs defects. \\

\addlinespace[5pt]
\multicolumn{2}{@{}l}{\textit{Text-only generation}} \\
\addlinespace[3pt]
Text-to-project &
Guides requirement analysis, implementation,
and validation. \\
\bottomrule
\end{tabularx}
\end{table}

\end{document}